\documentclass{article}

\PassOptionsToPackage{numbers, compress}{natbib}

\usepackage[preprint]{neurips_2026}

\usepackage[utf8]{inputenc} 
\usepackage[T1]{fontenc}    
\usepackage{hyperref}       
\usepackage{url}            
\usepackage{booktabs}       
\usepackage{amsfonts}       
\usepackage{nicefrac}       
\usepackage{microtype}      
\usepackage{xcolor}         
\usepackage{scalerel}
\usepackage{tikz}
\usepackage{wrapfig}

\usepackage{graphicx}
\usepackage{doi}
\usepackage{caption}
\usepackage{subcaption}
\usepackage{amsmath}
\usepackage{amsthm}
\usepackage{verbatim}
\usepackage{multirow}

\title{
Concept Graph Convolutions: Message Passing in the Concept Space
}

\author{
Lucie Charlotte Magister \\
  University of Cambridge\\
  Cambridge, UK\\
  \texttt{lcm67@cam.ac.uk} \\
  \And
  Pietro Li\`{o} \\
  University of Cambridge\\
  Cambridge, UK\\
  \texttt{pl219@cam.ac.uk} \\
}

\begin{document}

\maketitle

\begin{abstract}
The trust in the predictions of Graph Neural Networks is limited by their opaque reasoning process. Prior methods have tried to explain graph networks via concept-based explanations extracted from the latent representations obtained after message passing. However, these explanations fall short of explaining the message passing process itself. To this aim, we propose the Concept Graph Convolution, the first graph convolution designed to operate on node-level concepts for improved interpretability. The proposed convolutional layer performs message passing on a combination of raw and concept representations using structural and attention-based edge weights. We also propose a pure variant of the convolution, only operating in the concept space. Our results show that the Concept Graph Convolution allows to obtain competitive task accuracy, while enabling an increased insight into the evolution of concepts across convolutional steps.
\end{abstract}

\section{Introduction}

The opaque reasoning process of Graph Neural Networks (GNN, \citep{scarselli2008graph}) impedes human trust in their predictions \citep{graphxai_survey}, motivating a line of research focused on increasing their explainability and interpretability \citep{graphxai_survey, corso2024graph, ji2025comprehensive}. A subset of this research has focused on extracting concept-based explanations from the latent space of GNNs \citep{magister2021gcexplainer, magister2023concept, azzolin2023globalexplainabilitygnnslogic}, based on the natural clustering observed in the node representations after message passing \citep{magister2021gcexplainer}. While these explanations provide an insight into the latent space after message passing, they fail to explain the reasoning process performed during message passing itself. Specifically, what makes message passing unintepretable is the representation of nodes, computed via neighbourhood aggregation \citep{geng2026beyond}. The simple averaging of surrounding neighbours obscures the meaning of the nodes' representations. While attention-based graph convolutions \citep{brody2021attentive, miao2022interpretable, velickovic2018graph} improve this by weighting the contribution of neighbouring nodes based on attention, the node vector itself still remains uninterpretable. For a full understanding of GNNs, it is therefore desirable for a graph convolutional layer to operate on human interpretable concepts.

\begin{figure}
\centerline{\includegraphics[width=0.9\textwidth, trim=0.5cm 0cm 0.5cm 0cm, clip]{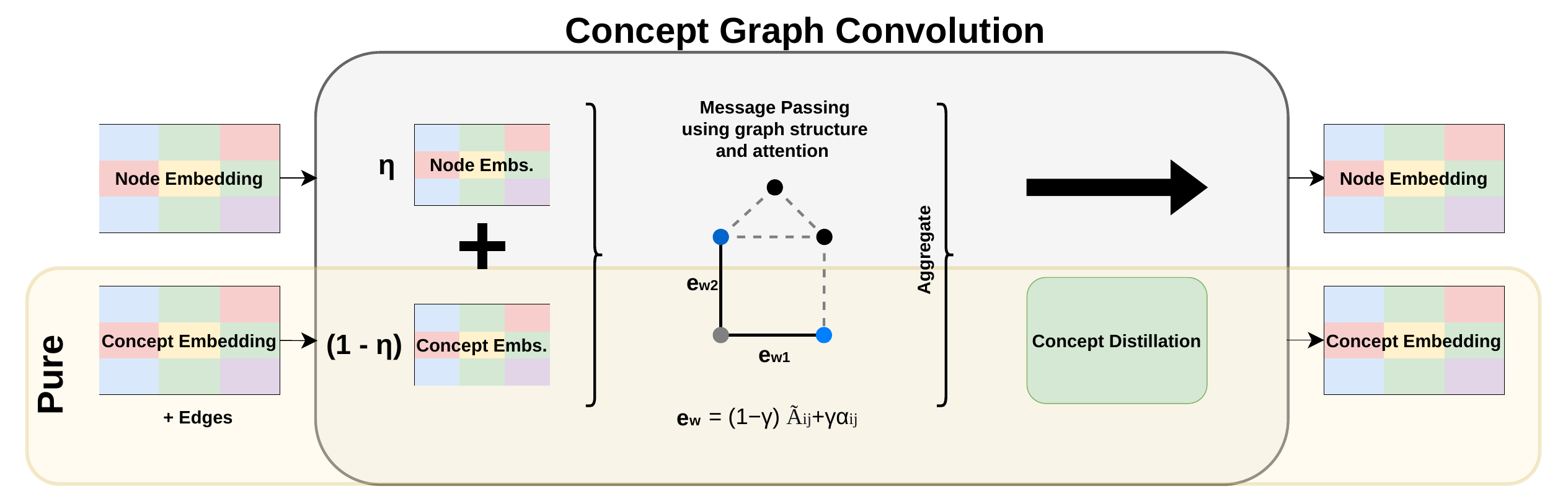}}
\caption{Visualisation of the processing steps of the CGC. The graph convolution combines raw and concept embeddings via a learned weighted scalar and then performs message passing, where edge weights are computed from the graph structure and attention. Finally, the convolution distils a new concept representation. The pure variant of the CGC does not rely on node embeddings but simply uses concept embeddings.}
\label{visual_abstract}
\end{figure}

To this aim, we introduce the Concept Graph Convolution (CGC). To the best of our knowledge, the CGC is the first graph convolution designed around operating on node-level concepts for improved interpretability. The CGC achieves this by representing nodes via both raw latent representations and concept representations. Both representations share a single linear transformation, projecting them into the same space. The learned scalar $\eta$ is then used to mix the representations and obtain a message for neighbourhood aggregation. Edge weights for neighbourhood aggregation are derived based on the graph structure and attention between the concept representations of two neighbouring nodes, inspired by the attention mechanism \citep{velickovic2018graph}. A learnable scalar $\gamma$ mixes the concept-based attention weights with the standard structural GCN weights. Finally, we obtain a valid concept representation again after message passing by applying a normalised softmax function to the resulting embedding. This design allows the CGC to perform message passing over both raw latent representations and concept representations for improved interpretability. We also propose the pure CGC, which discards the raw latent representations and purely performs message passing in the concept space. Most notably, the CGC layer provides interpretability through three distinct design features: (i) message passing in the concept space, (ii) a learned scalar explaining the importance of graph structure versus neighbouring concepts and (iii) concept attention. Figure \ref{visual_abstract} visually summarizes the CGC and the pure CGC variant.

The remainder of this paper is structured as follows. In Section \ref{related_work}, we review the related literature before introducing the Concept Graph Convolution in Section \ref{method}. We then detail our experimental setup and present experimental results in Sections \ref{experiments} and \ref{results}, respectively. Finally, we conclude with a discussion and conclusion in Sections \ref{discussion} and \ref{conclusion}.

\section{Related Work}
\label{related_work}

\paragraph{Graph convolutions} A number of prominent graph convolutional layers exist \citep{Kipf2016GCNConv, graphsagepaper, xu2018GINConv, chebconvpaper, velickovic2018graph, brody2021attentive}. The most prominent graph convolutional layer is the layer introduced in the Graph Convolution Network (GCN, \citep{Kipf2016GCNConv}), which performs a weighted sum of the features of neighbouring nodes based on the graph structure. A number of graph convolutions can be deemed to make message passing more interpretable. For example, the layer introduced in the Graph Attention Network (GAT, \citep{velickovic2018graph}) proposes to use attention to learn importance weights over neighbouring nodes. Transformed node features of a source node and neighbouring node are concatenated and passed through a feed-forward layer to get a raw attention score, which is then normalized across all neighbours of a node to obtain an attention score reflecting relative importance. Note, that the attention score is static, which means that the ranking of neighbours is independent of the source node. In contrast, the GATv2 \citep{brody2021attentive} modifies the original architecture of the GAT to use dynamic attention for improved expressivity. Dynamic attention allows a relative ranking of neighbours by changing the ordering of the attention operations. The Attention-based Graph Neural Network (AGNN, \citep{thekumparampil2018attention}) also uses dynamic attention implemented via cosine similarity, however, is limited in expressivity. In contrast, the Self-Supervised Graph Attention Network (SuperGAT, \citep{kim2022find}) learns attention scores via learning to predict edges in a self-supervised manner. This strongly aligns attention weights with the graph structure, making them more interpretable. Graph Stochastic Attention (GSAT, \citep{miao2022interpretable}) directly aims to improve interpretability by injecting random noise into the attention weights of a GNN to learn salient subgraph structures. In contrast, the Simple Graph Convolution (SGC, \citep{wu2019simplifying}) removes non-linearities between GCN layers and only applies a single softmax function at the end. While the simplified design makes feature attribution easier, the SGC suffers from oversmoothing and limited expressivity \citep{yang2020revisiting}. In contrast, the recent work SYMGraph \citep{geng2026beyond} directly aims to make graph convolutions more interpretable by replacing message passing with discrete structural hashing and topological role-based aggregation. While the aforementioned works increase transparency to a certain extent, they fail to make the hidden representations of nodes more interpretable, which we tackle with CGCs by operating in the concept space.

\paragraph{Message passing in the concept space} Only two other works have focused on performing message passing directly on concepts, namely Relational Concept Bottleneck Models (R-CBMs, \citep{barbiero2024relational}) and Causal Concept Graph Models (Causal CGMs, \citep{causal_concept_models}). R-CBMs map input entities into the concept embedding space and then perform message passing on the structure of the graph, after which a final prediction is extracted. While R-CBMs operate in the concept space, the goal of R-CBMs is slightly different. R-CBMs are a full relational architecture following the concept bottleneck principle, which means that the concepts extracted are directly used for prediction. While this improves the transparency of the interaction between concepts, it does not explain how the concepts arise. In contrast, CGCs are a convolutional layer architecture designed to trace the evolution and interplay of concepts. Causal CGMs \citep{causal_concept_models} extend the notion of R-CBMs by not assuming that concepts are independent, but rather causal. The graph convolutional layers in the model operate on nodes representing concepts and edges representing causal relationships. Operating in the concept representation space is a parallel of these two works and our work. However, there is one crucial difference. Both R-CBMs and Causal CGMs perform message passing on already embedded concepts and form a full model architecture. In contrast, CGC is a graph convolutional layer which extracts these concept representations as part of message passing, tackling the challenge of making graph convolutions more transparent. To the best of our knowledge, CGCs are the first graph convolutional layer to operate in the concept space and extract concepts at every convolutional step.

\paragraph{Attention as explanation} A number of aforementioned graph convolutional works rely on the attention mechanism to increase the interpretability of GNNs \citep{brody2021attentive, thekumparampil2018attention, velickovic2018graph}. This is based on attention weights providing an intuitive signal of which elements a model uses for prediction. However, it is contested whether attention actually provides valid explanations \citep{bibal2022attention, jain2019attention}. In particular, \citet{jain2019attention} find that attention is often not correlated with feature importance scores determined by gradient-based techniques on natural language processing (NLP) tasks. They show that varying attention masks can lead to the same prediction. In contrast, \citet{wiegreffe2019attention} argue that whether attention can be seen as a form of explanation depends on the evaluation criteria, such as comparing model performance when using uniform attention or comparing the attention maps learned by the model when trained with different seeds. \citet{bai2021attentions} approach the debate from a different angle. They argue that attention weights encode information beyond importance, which makes it unclear how to interpret them. To mitigate this issue, they propose random attention pretraining and instance weighting for mask-neutral learning. In contrast, \citet{akula2022attention} perform a human study to investigate how useful attention is as an explanation mechanism, specifically focusing on whether it increases human trust. They find that attention does not increase human trust compared to other non-attention baselines. However, the paper does not investigate whether attention faithfully reflects the model's underlying computation, a much stronger requirement for explanation. In conclusion, it is debated whether attention can be considered as explanation and we therefore must treat it with care, considering its limits and context in the design of the CGC.

\section{Concept Graph Convolution}
\label{method}

We propose the Concept Graph Convolution (CGC), the first interpretable graph convolution designed to incorporate node concept representations. At a high-level, the CGC works across both raw and concept representations. It first projects the raw and concept representations into the same latent space via a shared linear transformation. The resulting embeddings are then combined via a learned scalar $\eta$ for neighbourhood aggregation. For this, the CGC uses the standard GCN edge weights \citep{Kipf2016GCNConv} to discover structural properties, as well as single-headed attention between neighbouring concept representations, inspired by the interpretability of the GAT \citep{velickovic2018graph}. The layer combines these weights via the learnable scalar $\gamma$. To also obtain valid concept representations after neighbourhood aggregation, the CGC reapplies a normalised softmax function, as proposed in Concept Graph Networks (CGNs, \citep{magister2023concept}). This design allows increased insight into the evolution of concepts and the influence of neighbouring concepts on one another throughout a block of graph convolutional layers. We also propose the pure CGC, which discards raw latent representations and only considers concept representations in message passing. For ease of explanation, we will first discuss how we compute the edge weights, before describing how we perform message passing in the concept space.

\paragraph{Computing edge weights} An integral part of the interpretability of the CGC is the computation of edge weights. We compute two types of edge weights based on experimental observations. First, we compute a set of structural weights to capture the connectivity of the graph. We use the normalized adjacency weights for this, as proposed by \citet{Kipf2016GCNConv}: $\tilde{A}_{ij} = \frac{A_{ij}}{\sqrt{d_i d_j}}$, where $A_{ij}$ is the adjacency matrix with self-loops and $d_i$ is the degree of node $i$. Using these structural edge weights prevents from feature scaling due to node degree.

In parallel, we also compute attention-based edge weights to capture the interplay between concepts. Inspired by the GAT \citep{velickovic2018graph}, we compute the attention weights as: $\alpha_{ij} = \text{softmax}_j(e_{ij})$ where $e_{ij} = \text{LeakyReLU}\left(\mathbf{a}^\top [\mathbf{q}_i \,\|\, \mathbf{q}_j]\right)$, and $\mathbf{q}_i$ and $\mathbf{q}_j$ are the fuzzy concept encoding vectors of nodes $i$ and $j$, obtained via a scaled softmax function \citep{magister2023concept}. Note, that the main difference to attention in GATs \citep{velickovic2018graph} is that we directly operate on concept vectors in order to capture semantic similarity. We also restrict ourselves to single-headed attention as this is the most interpretable. We do not use dynamic attention as proposed by \citet{brody2021attentive}, because static attention is more easily interpretable. Static attention globally ranks the importance of nodes, while dynamic attention is based on the source node. While this makes dynamic attention more expressive, it also makes the computation more complex. 

We then combine the structural and attention-based edge weights in the following way: $w_{ij} = (1 - \gamma)\tilde{A}_{ij} + \gamma \alpha_{ij}$, where $\gamma \in (0, 1)$ is a learned scalar parameter, which controls the trade-off between structural and attention weights. We opt for combining the two types of edge weights because preliminary experiments using pure attention on node concept vectors did not capture graph structure effectively. We make the active design choice to make $\gamma$ a learnable parameter in order to not restrict the network's capacity. However, we do encourage attention to be used as much as possible through the following $L2$ loss term \citep{mcdonald2009ridge}: $\mathcal{L}_{\text{gamma}} = \lambda_a \sum_{\ell} (\gamma - \gamma_0)^2$, where $\gamma_0$ is a target value and $\lambda_a$ controls how strongly the network is regularized. In our experiments, we set the target $\gamma_0$ to 0.8 to encourage reliance on attention-based propagation. The motivation for this is that we want to learn which concepts are important for one another.

\paragraph{Concept-based message passing} Given the combined edge weights, we then perform message passing between nodes. Note, that in the CGC, we have two types of node representations: (1) the raw latent representation denoted by $\mathbf{u}_i^{l}$ and (2) a fuzzy concept embedding vector, which we denote by $\mathbf{q}_i^{l}$, where $i$ is a node and $l$ is the current layer. We first perform a shared linear transformation on the two types of representations for a node: $h_{ui}^{l} = \mathbf{u}_i^{l} W^{l}$ and $h_{qi}^{l} = \mathbf{q}_i^{l} W^{l}$, where $W^{l}$ is a learnable weight matrix, and $h_{ui}^{l}$ and $h_{qi}^{l}$ are the resulting node embeddings projected into a shared space. We then combine the two embeddings in the following way: $h_{i}^{l} = (1 - \eta)h_{ui}^{l} + \eta h_{qi}^{l}$, where $\eta$ is a learned scalar. Notice, that $h_{qi}^{l}$ is not in the concept space anymore, as we have linearly transformed it. However, applying this linear transformation allows us to combine both raw and concept representations and improves the model's expressivity. Similar to our regularization of $\gamma$, we apply a $L2$ loss on $\eta$: $\mathcal{L}_{\text{eta}} = \lambda_b \sum_{\ell} (\eta - \eta_0)^2$, where $\eta_0$ is a target value and $\lambda_b$ controls how strongly the network is regularized. In our experiments, we also set the target $\eta_0$ to 0.8 to encourage reliance on the concept representations. We then perform message passing on the enriched representation. 

Using the mixed latent embedding $h_i^{l}$, we then compute the message for message passing between nodes $i$ and $j$ by scaling $h_i^{l}$ by the computed edge importance \citep{bronstein2021geometricdeeplearninggrids}: $m_{ij}^{l} = w_{ij}^{l} \, h_i^{l}$, where $w_{ij}^{l}$ is the combined structural and attention-based edge weights and $m_{ij}^{l}$ the resulting message from node $i$ to node $j$ at layer $l$. To obtain the updated representation for node $j$ after message passing, we perform a simple sum over the messages and add a learnt bias term: $z_j^{l+1} = \sum_{i \in \mathcal N(j)\cup\{j\}} \bigl(m_{ij}^{l}\bigr) + b^{l}$, where $z_j^{l+1}$ is the resulting node representation of node $j$ after aggregating across the neighbouring nodes, scaled by the edge weights computed. In this equation, $b^{l}$ is a learnt bias term, that provides a baseline activation. It can be interpreted as how important a concept is even if there is no evidence for the concept from neighbours. Furthermore, the bias term breaks symmetry and allows to prefer a concept globally to another one. It can be interpreted as a global prior over concepts, telling us whether a concept is preferred or dispreferred.

Finally, we cast the aggregated node representation $z_j^{l+1}$ back into the concept space by applying a normalized softmax function, as introduced by \citep{magister2023concept}, yielding $\mathbf{q}_j^{l+1}$. This allows us to operate in the concept space across graph convolutions and track the evolution of concepts. Note that only in the first layer the input to the CGC are not concept embeddings, but the raw node features.

\paragraph{Pure Concept Graph Convolution} We note that while combining raw latent representations and concept representations may improve the model's expressivity, it hampers interpretability to a certain extent. To this end, we propose the pure CGC. The pure CGC discards the raw latent representation and simply operates on the concept representation. We will demonstrate the performance of both layers.

\paragraph{Adapting the definition of concept completeness for graph classification} Concept completeness measures how representative the discovered set of concepts is for predicting the final task label \citep{yeh2020completeness}. \citet{yeh2020completeness} propose computing concept completeness by training a classifier, such as a decision tree \citep{breiman1984classification}, to predict the task label from a given concept encoding for the input instance. In the context of graph classification, \citet{magister2021gcexplainer} propose predicting the graph label from the node-level concept. However, graph classification requires reasoning across multiple nodes, which a node-level concept cannot represent well. To this aim, we propose a new computation of the concept completeness score for graph classification. Specifically, we propose to aggregate across the concepts discovered to represent the graph. As concept labels are a binarization of the fuzzy encoding vector, we propose representing a graph as a frequency vector of the associated concepts. An alternative design to a frequency vector is a one-hot encoding of the present concepts, however, some graph classification tasks may rely on the frequency of a certain concept being present, wherefore, a frequency vector is more suitable. 

\section{Experiments}
\label{experiments}

Our experiments focus on answering the following research questions: (i) what is the impact of our approach on the generalization error of a GNN?, (ii) is the identified concept set complete with respect to the task? and (iii) are the concepts discovered coherent? how does a node concept evolve across convolutional layers? We hypothesize that our approach can: (i) obtain on par task accuracy with respect to a standard GCN \citep{Kipf2016GCNConv} and GAT \citep{velickovic2018graph}; (ii) identify a complete set of concepts, (iii) extract ground truth graph concepts aligned with human expectations, and (iv) showcase the evolution of node concepts throughout a graph convolutional block.

\paragraph{Metrics} We measure the performance and interpretability of the CGC on two key metrics. To be able to answer our first research question, we measure model performance via classification accuracy. To answer our second research question on explanation performance, we compute the concept completeness score of individual layers. Lastly, we examine concept interpretability by qualitatively examining the concepts discovered and their evolution across convolutional layers. 

\paragraph{Datasets} We perform the experiments on datasets for node classification and graph classification. Specifically, we use the same set of node and graph classification datasets as GNNExplainer \citep{ying2019gnnexplainer}, as subsequent research establishes these as benchmarks \citep{luo2020parameterized,vu2020pgm,magister2021gcexplainer,azzolin2023globalexplainabilitygnnslogic,xuanyuan2023global}. However, we further expand the set of graph classification datasets used to allow for a more insightful qualitative comparison. For this, we rely on datasets proposed by \citet{longa2022explaining}. 

\paragraph{Node classification} We use the five synthetic node classification datasets proposed by \citet{ying2019gnnexplainer}. We select these datasets, because they have a ground truth motif encoded, allowing for a rich analysis of the concepts discovered. The first dataset is BA-Shapes, which consists of a Barab\'asi-Albert (BA) graph~\cite{barabasi1999emergence} to which $80$ house motifs and $70$ random edges are attached. The goal of the dataset is to identify a node belonging to one of four classes: (1) the base graph, (2) the top of a house, (3) the middle of a house and (4) the bottom of a house. The second dataset BA-Community extends this task by forming a union of two BA-Shapes graphs and asking to classify the node into one of eight classes, which represent the structural role and graph membership. Similarly, the BA-Grid dataset is also based on a BA graph~\cite{barabasi1999emergence}, but has $80$ 3-by-3 grid structures attached to it, with the similar goal of identifying the node belonging to the base graph or grid structure. In contrast, the Tree-Grid and Tree-Cycles datasets are formed by binary trees of depth 8 with 80 3-by-3 grid structures and 6 node cycle structures attached, respectively. Again, the classification task here is to identify nodes belonging to the base graph or the attached motif.

\paragraph{Graph classification} We also test the performance of the CGC on four synthetic graph classification tasks and two real-world graph classification tasks. The four synthetic graph classification datasets are the Grid, Grid-House, Stars and House-Colour dataset, introduced by \citet{longa2022explaining} for evaluating the explainability of GNNs. Inspired by the node classification datasets introduced by \citet{ying2019gnnexplainer}, these datasets encode ground truth motifs for analysis. The Grid dataset consists of 1000 BA graphs of which half have a 3-by-3 grid structure attached. The task is to predict whether the grid structure is present. In contrast, the Grid-House dataset attaches both grid and house structures, where the task is to predict whether one of the motifs or both motifs are present. The STARS benchmark is a collection of random graphs generated by the Erd\H{o}s--R\'enyi (ER) random graph model \citep{Erdos1984OnTE} with one to four star-shaped structures attached. The task is to count the number of star structures attached, with classes corresponding to 1, 2 or 3-4 stars being attached. Finally, the House-Colour dataset introduces node features. The dataset consists of BA base graphs to which one to three coloured house structures are attached. Here, a blue house indicates the positive class, while a green house indicates a negative class. The remaining houses have randomly coloured nodes. 

We also include two real-world datasets to evaluate the performance of the CGC on less structured graph classification data. For this, we employ the Mutagenicity \cite{morris2020tudataset} dataset, which is a collection of graphs representing mutagenic and non-mutagenic molecules. The task is to identify whether a molecule is mutagenic or not. The second real-world dataset is Reddit-Binary \cite{morris2020tudataset}, which is a collection of graphs representing Reddit discussion threads. Here a node represents a user and an edge the interaction between users. The challenge of evaluating these datasets is that there is no ground truth motif beyond the ring structure and nitrogen dioxide compound in Mutagenicity, and the star-like structure in Reddit-Binary, as suggested by \citet{ying2019gnnexplainer}. We refer the reader to Appendix \ref{dataset_stats} for statistics on the datasets, such as the number of graphs, average graph size and number of classes.

\paragraph{Baselines and setup}

To the best of our knowledge, the CGC is the first concept-based graph convolution of its kind. This makes it difficult to select an appropriate baseline. We choose to compare our approach with a standard GCN \citep{Kipf2016GCNConv} and a GAT \citep{velickovic2018graph} for two key reasons. First, this selection of baselines allows us to validate that our model achieves on par task accuracy. Second, the comparison to the standard GCN and GAT allows us to validate our design choices in using edge weights from both the graph structure and attention. We make one key modification to the GCN and GAT to allow for improved comparison. Namely, we insert a normalized softmax after the graph convolutional layers of the models, as proposed in CGNs \citep{magister2023concept}, to extract node-level concepts and allow comparing the networks. We maintain the same set of hyperparameters across models to allow fair comparison. We refer the reader to Appendix \ref{training} for a summary of the model hyperparameters, inspired by the settings used by \citet{magister2023concept} and \citet{longa2022explaining}. While R-CBMs \citep{barbiero2024relational} are a full relational architecture, we compare our approach to them because they operate in the concept space and provide a more recent, interpretable baseline. Note, that we do not compare our approach to interpretable models such as the ProtGNN \citep{zhang2022protgnn} or MEGAN \citep{teufel2022megan}, because they simply use standard graph convolutional layers and do not enlighten the graph convolutional process via concepts. We also do not compare to SYMGraph \citep{geng2026beyond} or the work of \citet{he2025explaining} analysing gradients, because the works were released after we conducted our experiments. Moreover, we do not compare to post-hoc explainability methods, such as GNNExplainer~\cite{ying2019gnnexplainer}, GLGExplainer~\cite{azzolin2023globalexplainabilitygnnslogic} or LeArn Removal-based Attribution (LARA, \citep{rong2025efficient}), as the methods are post-hoc and do not explain what individual graph convolutional layers have learned.

\section{Results}
\label{results}

\paragraph{Concept graph convolutions allow for networks as accurate as standard graph convolutions} The CGC layer allows to construct GNNs as accurate as standard graph convolutions, like the GCN \citep{Kipf2016GCNConv} and GAT \citep{velickovic2018graph} layers, as shown by our results in Table \ref{fig:accuracy}. Across both node and graph classification tasks, the model using the CGC achieves competitive task accuracy. Specifically, we find the CGC to outperform the GCN and GAT convolution on all synthetic datasets. The poor performance of the GAT may be attributed to all synthetic datasets, except for the House-Colour dataset, relying on structure alone. Node features are initialized to a vector of ones, which hampers the learning of the GAT as initially all node representations look alike. However, this validates our design of computing both structural and attention-based edge weights. We also find that mixing latent and concept representations is beneficial for task accuracy, as the models using the CGC layer outperform the models using the pure CGC layer on 7 out of 11 datasets. The pure CGC variant only significantly outperforms the CGC layer on the BA-Community dataset, achieving an accuracy of 82.69\% compared to 69.56\%. All models perform badly on the Grid-House dataset, with the GCN achieving the highest accuracy at 63\%. For brevity, we refer the reader to Appendix \ref{rcbms} for a comparison to R-CBMs. Overall, we observe that the CGC and pure CGC layer allow for models as performant as models relying on standard, non-interpretable graph convolutional layers. This shows that we can avoid the accuracy-interpretability trade-off \citep{CEM_zarlenga} too the extent of the interpretability provided by the CGC layer.

\begin{table*}[ht]
\centering
\resizebox{1\textwidth}{!}{
\begin{tabular}{lllll}
\hline & \multicolumn{3}{c}{\textbf{\begin{tabular}[c]{@{}c@{}}Model Accuracy (\%)\end{tabular}}} \\
                        & \multicolumn{1}{c}{\textbf{CGC}}                       & \multicolumn{1}{c}{\textbf{Pure CGC}}      & \multicolumn{1}{c}{\begin{tabular}[c]{@{}c@{}}\textbf{GCN}\end{tabular}} & \multicolumn{1}{c}{\begin{tabular}[c]{@{}c@{}}\textbf{GAT}\end{tabular}} \\ \hline
\textbf{BA-Shapes}   & \textbf{98.86 (98.37, 99.34)} & 98.71 (97.56, 99.87) & 96.86 (95.02, 98.70) & 47.14 (47.14, 47.14) \\
\textbf{BA-Community} & 69.56 (66.17, 72.95) & \textbf{82.69 (77.51, 87.88)} & 70.55 (62.65, 78.45) & 65.86 (57.28, 74.44)   \\
\textbf{BA-Grid} & \textbf{99.90 (99.63, 100.00)} & \textbf{99.90 (99.63, 100.00)} & 98.82 (97.90, 99.75)  & 71.27 (66.30, 76.25)  \\
\textbf{Tree-Cycle} & \textbf{97.69 (95.17, 100.00)} & 96.98 (93.80, 100.00) & 84.62 (76.94, 92.31) & 50.55 (46.45, 54.66)  \\
\textbf{Tree-Grid} & \textbf{99.43 (98.59, 100.00)} & 99.19 (98.25, 100.00) & 85.99 (78.75, 93.23) & 59.35 (56.04, 62.66)  \\
\hline
\textbf{Grid} & \textbf{99.15 (97.73, 100.00)} & 97.90 (97.01, 98.79) & 87.60 (85.57, 89.63) & 50.00 (50.00, 50.00)   \\
\textbf{Grid-House} & 55.40 (48.88, 61.92) & 55.50 (49.71, 61.29) & \textbf{63.00 (62.56, 63.44)} & 50.00 (50.00, 50.00) \\
\textbf{STARS} & \textbf{98.93 (98.75, 99.12)} & 98.80 (98.57, 99.03) & 97.80 (97.57, 98.03) & 33.33 (33.33, 33.33) \\
\textbf{House-Colour}  & 98.40 (96.73, 100.00) & 97.40 (95.37, 99.43) & \textbf{98.5 (97.62, 99.38)} & 98.20 (96.52, 99.88)\\
\hline
\textbf{Mutagenicity} & 77.40 (73.62, 81.17) & \textbf{77.67 (75.28, 80.06)} & 75.58 (73.17, 77.99) & 76.59 (73.91, 79.27) \\
\textbf{Reddit-Binary} & \textbf{87.55 (84.36, 90.74)} & 84.75 (80.42, 89.08) & 72.95 (70.43, 75.47) & 48.05 (45.67, 50.43) \\
\hline
\end{tabular}%
}
\caption{Model accuracy for models trained with Concept Graph Convolution (CGC), pure CGC, Graph Convolution Network (GCN, \citep{Kipf2016GCNConv}) and Graph Attention Network (GAT, \citep{velickovic2018graph}) layers.}
\label{fig:accuracy}
\end{table*}

\paragraph{Concept graph convolutions discover a complete set of node concepts} Table \ref{fig:concept_completenss} summarizes the concept completeness scores at the final convolutional layer of the models trained to allow comparison across architectures. The pure CGC is the best performing model on 4 out of 5 node classification tasks, achieving the highest completeness scores. Only the full CGC outperforms it on the BA-Shapes dataset. When we review the graph classification tasks, recall that we proposed a new computation of concept completeness for graph classification. The CGC, pure CGC and GCN perform similarly across the Grid, STARS and House-Colour datasets. All models have a poor concept completeness on the Grid-House dataset, as all models failed to learn the task well. The CGC, pure CGC and GCN also perform similarly in terms of concept completeness on the Mutagenicity and Reddit-Binary dataset. However, the concept completeness score at the final layer does not accurately capture the added benefit of the CGC and pure CGC, which is the increased model insight we gain across layers. Specifically, the CGC and pure CGC allow to also measure concept completeness at earlier layers of the model.

\begin{table*}[ht]
\centering
\resizebox{1\textwidth}{!}{
\begin{tabular}{lllll}
\hline & \multicolumn{3}{c}{\textbf{\begin{tabular}[c]{@{}c@{}}Concept Completeness (\%)\end{tabular}}} \\
                        & \multicolumn{1}{c}{\textbf{CGC}}                       & \multicolumn{1}{c}{\textbf{Pure CGC}}      & \multicolumn{1}{c}{\begin{tabular}[c]{@{}c@{}}\textbf{GCN}\end{tabular}} & \multicolumn{1}{c}{\begin{tabular}[c]{@{}c@{}}\textbf{GAT}\end{tabular}}      \\ \hline
\textbf{BA-Shapes}   & \textbf{96.71 (94.16, 99.27)} & 84.57 (78.94, 90.20) & 95.29 (90.44, 100.00) & 47.14 (47.14, 47.14) \\
\textbf{BA-Community} & 67.24 (63.64, 70.83)& \textbf{74.15 (68.83, 79.46)} & 60.99 (53.80, 68.18) & 62.79 (56.26, 69.32)  \\
\textbf{BA-Grid} & 99.41 (98.09, 100.00) & \textbf{100.00 (100.00, 100.00)} & 94.02 (90.31, 97.73) & 71.27 (66.30, 76.25)\\
\textbf{Tree-Cycle} & 92.26 (85.60, 98.93) & \textbf{95.68 (92.80, 98.56)} & 77.49 (57.53, 97.45) & 50.25 (46.09, 54.41)\\
\textbf{Tree-Grid} & 93.85 (87.94, 99.76) & \textbf{96.92 (93.32, 100.00)} & 83.97 (78.14, 89.79) & 59.35 (56.04, 62.66) \\
\hline
\textbf{Grid} & \textbf{99.05 (97.59, 100.00)} & 97.70 (96.47, 98.93) & 91.40 (82.14, 100.00) & 49.75 (49.75, 49.75) \\
\textbf{Grid-House} & 50.00 (50.00, 50.00) & 50.00 (50.00, 50.00) &  \textbf{50.3 (49.47, 51.13)} & 50.00 (50.00, 50.00)\\
\textbf{STARS} & \textbf{98.47 (97.51, 99.42)} & 98.20 (96.63, 99.77) & 97.07 (95.53, 98.60) & 54.67 (54.67, 54.67) \\
\textbf{House-Colour}  & 97.10 (94.21, 99.99) & 96.40 (95.88, 96.92) & \textbf{97.8 (96.24, 99.36)} & 97.20 (94.89, 99.51) \\
\hline
\textbf{Mutagenicity} & 71.68 (68.78, 74.58) & 71.54 (70.12, 72.96) & \textbf{72.51 (68.92, 76.11)} & 71.36 (68.71, 74.01)
 \\
\textbf{Reddit-Binary} & \textbf{84.45 (81.93, 86.97)} & 80.90 (75.67, 86.14) & 79.6 (75.83, 83.37) & 68.95 (66.44, 71.46) \\
\hline
\end{tabular}%
}
\caption{Concept completeness at the final convolutional layer of models trained with Concept Graph Convolution (CGC), pure CGC, Graph Convolution Network (GCN, \citep{Kipf2016GCNConv}) and Graph Attention Network (GAT, \citep{velickovic2018graph}) layers.}
\label{fig:concept_completenss}
\end{table*}

\begin{figure}[h!]
    \centering
    \subfloat[Concept Completeness Score over \textbf{CGC} Layers for Node Classification]{
        \includegraphics[width=0.47\textwidth, trim=0.25cm 0.4cm 0.25cm 1.5cm, clip]{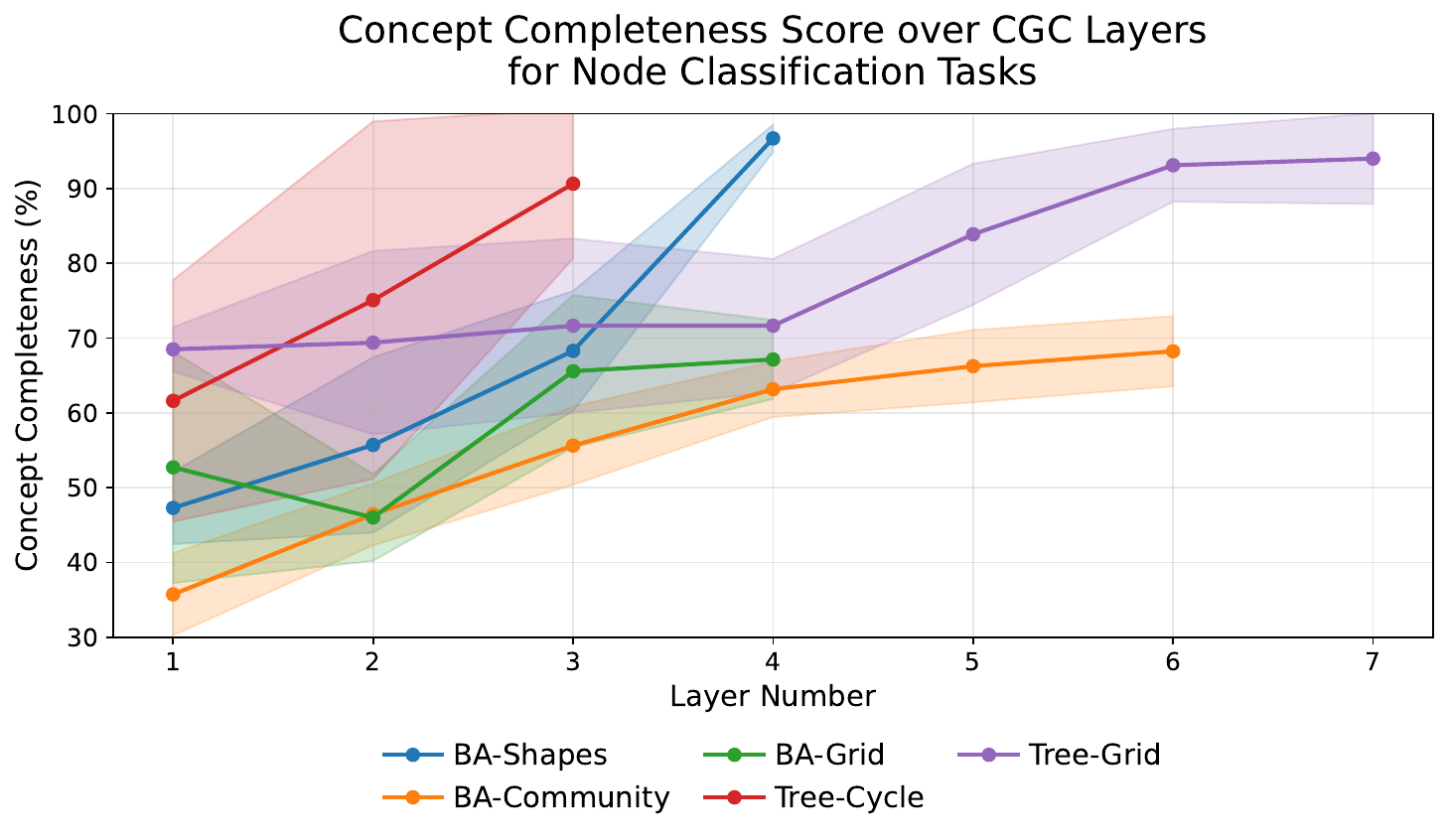}
        \label{fig:plot1_comp}
    }
    \hfill
    \subfloat[Concept Completeness Score over \textbf{Pure CGC} Layers for Node Classification]{
        \includegraphics[width=0.47\textwidth, trim=0.25cm 0.4cm 0.25cm 1.5cm, clip]{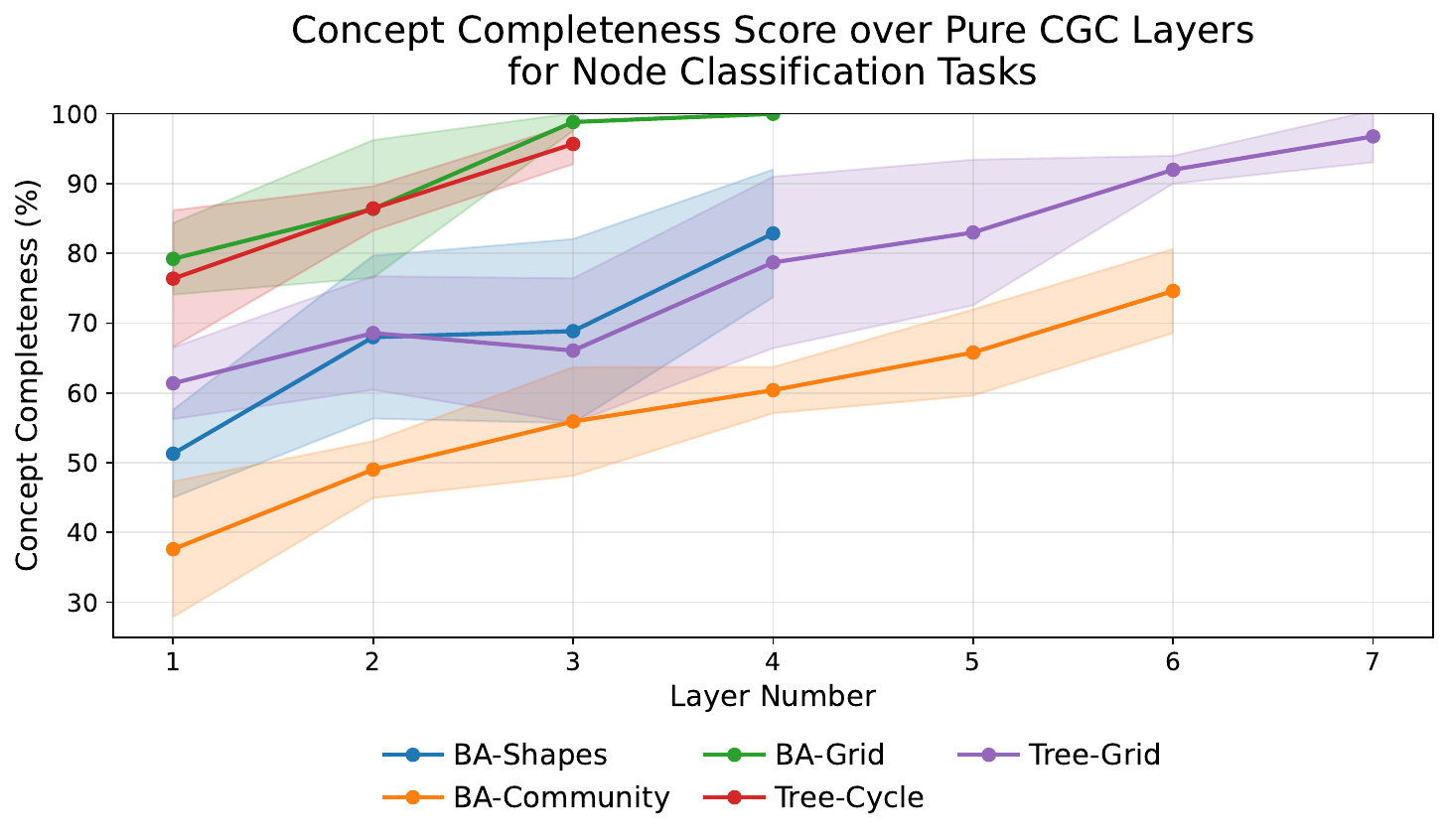}
        \label{fig:plot2_comp}
    }
    \caption{\raggedright Concept completeness measured across Concept Graph Convolution (CGC) and pure CGC layers trained on different node classification tasks. We can observe that concept completeness increases across layers in the model, indicating that the size of the receptive field matters to the task and produces improved concepts.}
    \label{fig:concept_completeness_plot_node}
\end{figure}

\paragraph{Concept graph convolutions allow to track concept evolution} We can track concept evolution over the CGC and pure CGC layers, as shown in Figures \ref{fig:concept_completeness_plot_node} and \ref{fig:concept_completeness_plot_graph}. In reference to Figure \ref{fig:concept_completeness_plot_node}, we see that the ability to discriminate the final node classification task via the concept representation increases with every layer in the network. This indicates that the increased receptive field of deeper layers is important for the task and allows us to distil better representations. For brevity, we refer the reader to Appendix \ref{graph_class_evolution} for the analysis of concept evolution in graph classification tasks. For brevity, we refer the reader to Appendix \ref{oversmoothing} and \ref{training_time} for an analysis of concepts with oversmoothing and over training time, respectively.

\captionsetup[subfloat]{justification=centering}

\begin{figure}[h!]
    \centering
    \subfloat[Gamma across layers for \textbf{CGC} layers in node classification models]{
        \includegraphics[width=0.48\textwidth, trim=0.25cm 0.4cm 0.25cm 1.8cm, clip]{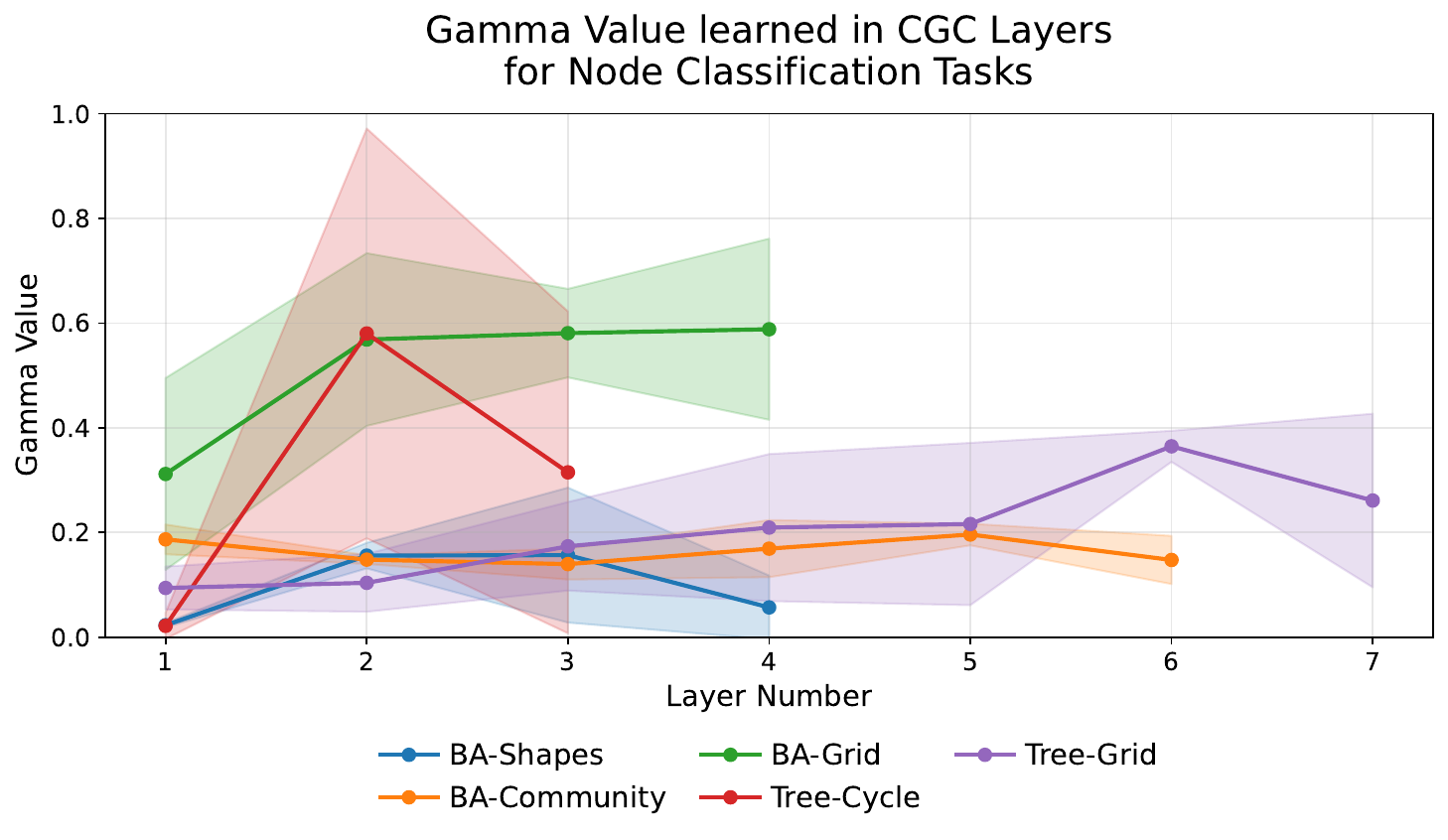}
        \label{fig:gamma_a}
    }
    \hfill
    \subfloat[Gamma across layers for \textbf{pure CGC} layers in node classification models]{
        \includegraphics[width=0.48\textwidth, trim=0.25cm 0.4cm 0.25cm 1.8cm, clip]{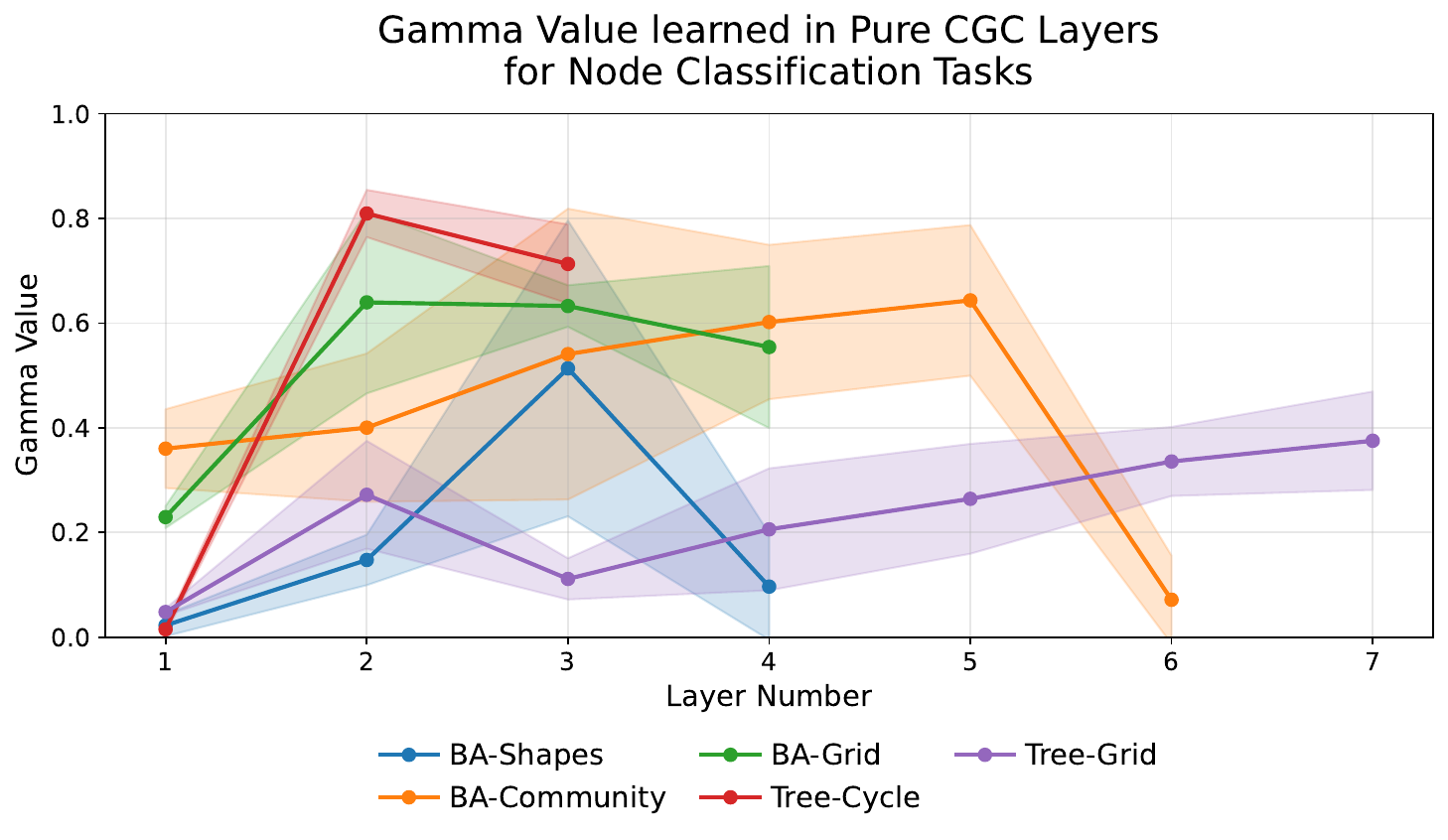}
        \label{fig:gamma_b}
    }
    \caption{\raggedright We plot the learned $\gamma$ value of Concept Graph Convolution (CGC) and pure CGC layers across the models trained for node classification.}
    \label{fig:gamma}
\end{figure}

\paragraph{Structure versus concept attention in message passing} The learned mixing parameter $\gamma$ determines the trade-off between structure versus concept attention in CGC layers. A small $\gamma$ value means that the model focuses mostly on neighbourhood structure, rather than attention scores for neighbouring concepts. Figure \ref{fig:gamma} shows the gamma value for different layers in the models trained across the node and graph classification datasets. In reference to Figure \ref{fig:gamma_a}, we observe fairly low $\gamma$ values across graph convolutional layers. Specifically, $\gamma < 0.4$ for most layers in the models trained on the BA-Shapes, BA-Community, Tree-Grid and Tree-Cycle datasets. In contrast, only the $\gamma$ value for the model trained on BA-Grid takes a value of $\sim0.60$ in the final CGC layer. The low values for $\gamma$ make intuitive sense, as the graph structure is the most important element for predicting the node classification tasks. For example, in the BA-Shapes dataset the task is to identify certain nodes in the house structure. All tasks except BA-Community are purely structural. BA-Community is also structural but relies on the node features to encode the node belonging to a certain community. Overall, it can be state that the value of $\gamma$ increases with deeper CGC layers, indicating that the model first pays attention to structure and then focuses more on node concepts. This is aligned with the observation that the set of concepts distilled in later layers is more representative and meaningful, wherefore attention on concepts is more meaningful. However, we do not observe the same trend for pure CGC layers, whose $\gamma$ values are depicted in Figure \ref{fig:gamma_b}. In contrast, the $\gamma$ values seem to climb more steeply in early layers but then decline in the last layer, except for the model trained on the Tree-Grid dataset. The steeper rise in the $\gamma$ value may stem from the pure CGC layer operating purely in the concept space, where simply aggregating the sparse vectors via structure may not create a meaningful signal and attention becomes more important until later layers, which may have more disentangled concept representations. The decline in the $\gamma$ value in the last pure CGC layers could indicate that a shallower model would be more desirable here. This is in line with our observation that a shallower network is more desirable for improved concept completeness in the BA-Shapes dataset in Appendix \ref{oversmoothing}. For brevity, we refer the reader to Appendix \ref{gamma} for an analysis of $\gamma$ across layers in graph classification models. We refer the reader to Appendix \ref{eta} for an analysis of CGC reliance on node embeddings versus concept embeddings.

\subsubsection{Visualisation of concept-based message passing}

\begin{figure}[h]
\centerline{\includegraphics[width=0.85\textwidth]{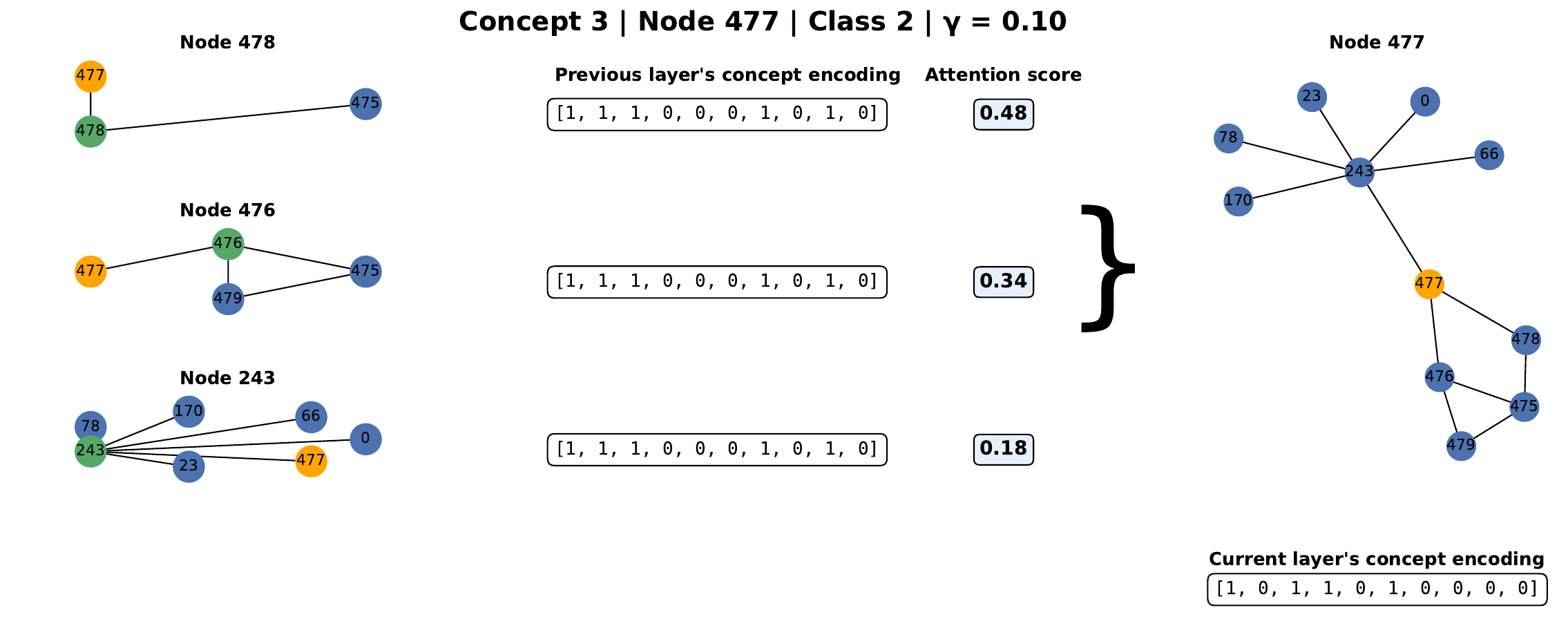}}
\caption{A concept discovered for the BA-Shapes dataset at the second pure Concept Graph Convolution (CGC) layer of the model. The graph on the left visualises the concept discovered, while the graphs on the right are the expanded graphs of the adjacent nodes in the previous layer. The orange node is the node assigned to the concept, while the green nodes are the respective neighbour in question and the blue nodes represent the other neighbours.}
\label{att_fig1}
\end{figure}

Finally, we can provide a deeper insight into the model via more elaborate visualizations of how concepts arise. We can visualize the concept found at a given layer and how the concepts are the previous layer contribute to its computation. Figure \ref{att_fig1} visualizes the concept representing a bottom-of-the-house node in the BA-Shapes dataset found in the second pure CGC layer. Three nodes contribute to the computation of the new concept representation for the clustered node shown in orange. Specifically, the two adjacent nodes part of the house-structure are considered important, receiving an attention score of 0.48 and 0.34, respectively. In comparison, the third node, part of the base structure, receives a lower attention score at 0.18. This shows that the GNN correctly pays attention to neighbouring concepts that identify the house structure. When we examine the concept encoding vector, we can see that all three nodes are assigned the same concept. While this is correct for the first two nodes (nodes 478 and 476) as they are both part of the house structure, node 243 should not have the same concept vector assigned. However, this is aligned with the concept completeness score at second layer for the model trained on the BA-Shapes dataset (Figure \ref{fig:plot2_comp}) only being $\sim78\%$, which means the set of concepts is not complete and requires further disentanglement via more unique concept representations. We also must consider the value of $\gamma$ that indicates that the message constructed based on attention is only mixed into the message computed based on structure with $\gamma = 0.10$. While the assigned attention scores are reasonable, all three concept encodings are the same, thus only aggregating over this would not improve our concept representation, wherefore, the structural bias is needed to also incorporate other nodes which may change the concept encoding. Moreover, we must consider that the low focus on the attention scores indicating by $\gamma$ may mean that the scores are not well calibrated and should be interpreted with care. Overall, the design of the pure CGC layer provides an improved ability to trace the reasoning process of the GNN, making GNNs more interpretable by design. We refer the reader to Appendix \ref{vis} for further concept visualisations.

\section{Discussion}
\label{discussion}
The CGC is the first concept-based graph convolution, allowing to monitor the refinement of concepts as the receptive field of the computation for the node state grows. A core limitation of the SCN is that it repeatedly uses a scaled softmax function \citep{magister2023concept} to encourage clustering in the model's latent space. While this is a simple solution to encouraging clustering, it can lead to information loss and vanishing gradients as the embeddings are repeatedly scaled. This drawback is an interesting area for future work. Nevertheless, in our experiments we do not find the repeated use of the softmax function to be an issue. A strength of the CGC is its simplicity: a slight modification of the traditional GCN \citep{Kipf2016GCNConv} architecture has allowed a deeper insight into the processing of the GNN.

\section{Conclusion}
\label{conclusion}

We show that CGCs allow us to successfully follow the evolution of concepts across layers. Specifically, we show that CGC layers do not deteriorate task accuracy, while providing an improved insight into the reasoning process of GNNs. We show that a main benefit of CGCs is that we can track concept evolution across multiple layers, providing subgraph explanations at different levels of granularity. Moreover, we find that the pure CGC performs similarly in regard to task accuracy compared to the relaxed variant. It can be argued that a model using pure CGC layers is more interpretable, as it forces the reasoning into the concept space. In contrast, the relaxed CGC layer also incorporates raw latent embeddings into the computation, slightly improving task accuracy. However, it still discerns the model's reliance on the less restricted representations over the concept representations. Overall, the CGC allows a deeper analysis of a model's reasoning process compared to standard graph convolutions, while maintaining predictive accuracy.

\bibliographystyle{unsrtnat}
\bibliography{references}

@article{paszke2019pytorch,
  title={Pytorch: An imperative style, high-performance deep learning library},
  author={Paszke, Adam and Gross, Sam and Massa, Francisco and Lerer, Adam and Bradbury, James and Chanan, Gregory and Killeen, Trevor and Lin, Zeming and Gimelshein, Natalia and Antiga, Luca and others},
  journal={arXiv preprint arXiv:1912.01703},
  year={2019}
}

@article{magister2021gcexplainer,
  title={GCExplainer: Human-in-the-Loop Concept-based Explanations for Graph Neural Networks},
  author={Magister, Lucie Charlotte and Kazhdan, Dmitry and Singh, Vikash and Li{\`o}, Pietro},
  journal={arXiv preprint arXiv:2107.11889},
  year={2021}
}

@article{ying2019gnnexplainer,
  title={Gnnexplainer: Generating explanations for graph neural networks},
  author={Ying, Zhitao and Bourgeois, Dylan and You, Jiaxuan and Zitnik, Marinka and Leskovec, Jure},
  journal={Advances in neural information processing systems},
  volume={32},
  year={2019}
}

@article{luo2020parameterized,
  title={Parameterized explainer for graph neural network},
  author={Luo, Dongsheng and Cheng, Wei and Xu, Dongkuan and Yu, Wenchao and Zong, Bo and Chen, Haifeng and Zhang, Xiang},
  journal={Advances in neural information processing systems},
  volume={33},
  pages={19620--19631},
  year={2020}
}

@article{vu2020pgm,
  title={Pgm-explainer: Probabilistic graphical model explanations for graph neural networks},
  author={Vu, Minh and Thai, My T},
  journal={Advances in neural information processing systems},
  volume={33},
  pages={12225--12235},
  year={2020}
}

@article{morris2020tudataset,
  title={Tudataset: A collection of benchmark datasets for learning with graphs},
  author={Morris, Christopher and Kriege, Nils M and Bause, Franka and Kersting, Kristian and Mutzel, Petra and Neumann, Marion},
  journal={arXiv preprint arXiv:2007.08663},
  year={2020}
}

@article{scarselli2008graph,
  title={The graph neural network model},
  author={Scarselli, Franco and Gori, Marco and Tsoi, Ah Chung and Hagenbuchner, Markus and Monfardini, Gabriele},
  journal={IEEE transactions on neural networks},
  volume={20},
  number={1},
  pages={61--80},
  year={2008},
  publisher={IEEE}
}

@article{barabasi1999emergence,
  title={Emergence of scaling in random networks},
  author={Barab{\'a}si, Albert-L{\'a}szl{\'o} and Albert, R{\'e}ka},
  journal={science},
  volume={286},
  number={5439},
  pages={509--512},
  year={1999},
  publisher={American Association for the Advancement of Science}
}

@inproceedings{Kipf2016GCNConv,
author = {Kipf, Thomas N. and Welling, Max},
booktitle = {5th International Conference on Learning Representations, ICLR 2017 - Conference Track Proceedings},
eprint = {1609.02907},
month = {sep},
pages = {11313--11320},
publisher = {International Conference on Learning Representations, ICLR},
title = {{Semi-Supervised Classification with Graph Convolutional Networks}},
url = {http://arxiv.org/abs/1609.02907},
year = {2016}
}

@article{xu2018GINConv,
  title={How powerful are graph neural networks?},
  author={Xu, Keyulu and Hu, Weihua and Leskovec, Jure and Jegelka, Stefanie},
  journal={arXiv preprint arXiv:1810.00826},
  year={2018}
}

@book{breiman1984classification,
  title={Classification and Regression Trees},
  author={Breiman, L and Friedman, JH and Olshen, R and Stone, CJ},
  year={1984},
  publisher={Wadsworth}
}

@inproceedings{zhang2022protgnn,
  title={Protgnn: Towards self-explaining graph neural networks},
  author={Zhang, Zaixi and Liu, Qi and Wang, Hao and Lu, Chengqiang and Lee, Cheekong},
  booktitle={Proceedings of the AAAI Conference on Artificial Intelligence},
  volume={36},
  pages={9127--9135},
  year={2022}
}

@article{teufel2022megan,
  title={MEGAN: Multi-Explanation Graph Attention Network},
  author={Teufel, Jonas and Torresi, Luca and Reiser, Patrick and Friederich, Pascal},
  journal={arXiv preprint arXiv:2211.13236},
  year={2022}
}

@inproceedings{yeh2020completeness,
author = {Yeh, Chih-Kuan and Kim, Been and Arık, Sercan {\"{O}} and Li, Chun-Liang and Pfister, Tomas and Ravikumar, Pradeep},
booktitle = {Advances in Neural Information Processing Systems 33 (NeurIPS 2020)},
pages = {20554--20565},
title = {{On Completeness-aware Concept-Based Explanations in Deep Neural Networks}},
url = {https://papers.nips.cc/paper/2020/hash/ecb287ff763c169694f682af52c1f309-Abstract.html},
volume = {33},
year = {2020}
}

@misc{bronstein2021geometricdeeplearninggrids,
      title={Geometric Deep Learning: Grids, Groups, Graphs, Geodesics, and Gauges}, 
      author={Michael M. Bronstein and Joan Bruna and Taco Cohen and Petar Veličković},
      year={2021},
      eprint={2104.13478},
      archivePrefix={arXiv},
      primaryClass={cs.LG},
      url={https://arxiv.org/abs/2104.13478}, 
}

@inproceedings{azzolin2023globalexplainabilitygnnslogic,
title={Global Explainability of GNNs via Logic Combination of Learned Concepts}, 
author={Steve Azzolin and Antonio Longa and Pietro Barbiero and Pietro Liò and Andrea Passerini},
booktitle={International Conference on Learning Representations},
year={2023},
organization={PMLR}
}

@article{velickovic2018graph,
  title="{Graph Attention Networks}",
  author={Veli{\v{c}}kovi{\'{c}}, Petar and Cucurull, Guillem and Casanova, Arantxa and Romero, Adriana and Li{\`{o}}, Pietro and Bengio, Yoshua},
  journal={International Conference on Learning Representations},
  year={2018},
  url={https://openreview.net/forum?id=rJXMpikCZ},
}

@inproceedings{chebconvpaper,
author = {Defferrard, Micha\"{e}l and Bresson, Xavier and Vandergheynst, Pierre},
title = {Convolutional neural networks on graphs with fast localized spectral filtering},
year = {2016},
isbn = {9781510838819},
publisher = {Curran Associates Inc.},
address = {Red Hook, NY, USA},
booktitle = {Proceedings of the 30th International Conference on Neural Information Processing Systems},
pages = {3844–3852},
numpages = {9},
location = {Barcelona, Spain},
series = {NIPS'16}
}

@inproceedings{graphsagepaper,
author = {Hamilton, William L. and Ying, Rex and Leskovec, Jure},
title = {Inductive representation learning on large graphs},
year = {2017},
isbn = {9781510860964},
publisher = {Curran Associates Inc.},
address = {Red Hook, NY, USA},
booktitle = {Proceedings of the 31st International Conference on Neural Information Processing Systems},
pages = {1025–1035},
numpages = {11},
location = {Long Beach, California, USA},
series = {NIPS'17}
}

@InProceedings{pmlr-v70-li17f,
  title = 	 {Stochastic Modified Equations and Adaptive Stochastic Gradient Algorithms},
  author =       {Qianxiao Li and Cheng Tai and Weinan E},
  booktitle = 	 {Proceedings of the 34th International Conference on Machine Learning},
  pages = 	 {2101--2110},
  year = 	 {2017},
  editor = 	 {Precup, Doina and Teh, Yee Whye},
  volume = 	 {70},
  series = 	 {Proceedings of Machine Learning Research},
  month = 	 {06--11 Aug},
  publisher =    {PMLR},
  url = 	 {https://proceedings.mlr.press/v70/li17f.html},
}

@inproceedings{class_activation_mapping,
  author={Zhou, Bolei and Khosla, Aditya and Lapedriza, Agata and Oliva, Aude and Torralba, Antonio},
  booktitle={2016 IEEE Conference on Computer Vision and Pattern Recognition (CVPR)}, 
  title={Learning Deep Features for Discriminative Localization}, 
  year={2016},
  pages={2921-2929},
  doi={10.1109/CVPR.2016.319}}

@INPROCEEDINGS{gradcam,
  author={Selvaraju, Ramprasaath R. and Cogswell, Michael and Das, Abhishek and Vedantam, Ramakrishna and Parikh, Devi and Batra, Dhruv},
  booktitle={2017 IEEE International Conference on Computer Vision (ICCV)}, 
  title={Grad-CAM: Visual Explanations from Deep Networks via Gradient-Based Localization}, 
  year={2017},
  pages={618-626},
  doi={10.1109/ICCV.2017.74}
}

@inproceedings{CEM_zarlenga, author = {Zarlenga, Mateo Espinosa and Barbiero, Pietro and Ciravegna, Gabriele and Marra, Giuseppe and Giannini, Francesco and Diligenti, Michelangelo and Shams, Zohreh and Precioso, Frederic and Melacci, Stefano and Weller, Adrian and Lio, Pietro and Jamnik, Mateja}, title = {Concept embedding models: beyond the accuracy-explainability trade-off}, year = {2024}, isbn = {9781713871088}, publisher = {Curran Associates Inc.}, address = {Red Hook, NY, USA}, booktitle = {Proceedings of the 36th International Conference on Neural Information Processing Systems}, articleno = {1555}, numpages = {14}, location = {New Orleans, LA, USA}, series = {NIPS '22} }

@article{longa2022explaining,
  title={Explaining the explainers in graph neural networks: a comparative study},
  author={Longa, Antonio and Azzolin, Steve and Santin, Gabriele and Cencetti, Giulia and Lio, Pietro and Lepri, Bruno and Passerini, Andrea},
  journal={ACM Computing Surveys},
  volume={57},
  number={5},
  pages={1--37},
  year={2025},
  publisher={ACM New York, NY}
}

@article{Erdos1984OnTE,
  title={On the evolution of random graphs},
  author={Paul L. Erdos and Alfr{\'e}d R{\'e}nyi},
  journal={Transactions of the American Mathematical Society},
  year={1984},
  volume={286},
  pages={257-257},
  url={https://api.semanticscholar.org/CorpusID:6829589}
}

@article{corso2024graph,
  title={Graph neural networks},
  author={Corso, Gabriele and Stark, Hannes and Jegelka, Stefanie and Jaakkola, Tommi and Barzilay, Regina},
  journal={Nature Reviews Methods Primers},
  volume={4},
  number={1},
  pages={17},
  year={2024},
  publisher={Nature Publishing Group UK London}
}

@article{ji2025comprehensive,
  title={A comprehensive survey on self-interpretable neural networks},
  author={Ji, Yang and Sun, Ying and Zhang, Yuting and Wang, Zhigaoyuan and Zhuang, Yuanxin and Gong, Zheng and Shen, Dazhong and Qin, Chuan and Zhu, Hengshu and Xiong, Hui},
  journal={Proceedings of the IEEE},
  year={2025},
  publisher={IEEE}
}

@article{barbiero2024relational,
  title={Relational concept bottleneck models},
  author={Barbiero, Pietro and Giannini, Francesco and Ciravegna, Gabriele and Diligenti, Michelangelo and Marra, Giuseppe},
  journal={Advances in Neural Information Processing Systems},
  volume={37},
  pages={77663--77685},
  year={2024}
}

@inproceedings{causal_concept_models,
  author       = {Gabriele Dominici and
                  Pietro Barbiero and
                  Mateo Espinosa Zarlenga and
                  Alberto Termine and
                  Martin Gjoreski and
                  Giuseppe Marra and
                  Marc Langheinrich},
  title        = {Causal Concept Graph Models: Beyond Causal Opacity in Deep Learning},
  booktitle    = {The Thirteenth International Conference on Learning Representations,
                  {ICLR} 2025, Singapore, April 24-28, 2025},
  year         = {2025},
}

@inproceedings{miao2022interpretable,
  title={Interpretable and generalizable graph learning via stochastic attention mechanism},
  author={Miao, Siqi and Liu, Mia and Li, Pan},
  booktitle={International conference on machine learning},
  pages={15524--15543},
  year={2022},
  organization={PMLR}
}

@article{graphxai_survey,
	author = {Nandan, Mauparna and Mitra, Soma and De, Debashis},
	isbn = {1433-3058},
	journal = {Neural Computing and Applications},
	number = {17},
	pages = {10949--11000},
	title = {GraphXAI: a survey of graph neural networks (GNNs) for explainable AI (XAI)},
	url = {https://doi.org/10.1007/s00521-025-11054-3},
	volume = {37},
	year = {2025},
}

@article{brody2021attentive,
  title={How attentive are graph attention networks?},
  author={Brody, Shaked and Alon, Uri and Yahav, Eran},
  journal={arXiv preprint arXiv:2105.14491},
  year={2021}
}

@article{thekumparampil2018attention,
  title={Attention-based graph neural network for semi-supervised learning},
  author={Thekumparampil, Kiran K and Wang, Chong and Oh, Sewoong and Li, Li-Jia},
  journal={arXiv preprint arXiv:1803.03735},
  year={2018}
}

@inproceedings{wu2019simplifying,
  title={Simplifying graph convolutional networks},
  author={Wu, Felix and Souza, Amauri and Zhang, Tianyi and Fifty, Christopher and Yu, Tao and Weinberger, Kilian},
  booktitle={International conference on machine learning},
  pages={6861--6871},
  year={2019},
  organization={Pmlr}
}

@article{kim2022find,
  title={How to find your friendly neighborhood: Graph attention design with self-supervision},
  author={Kim, Dongkwan and Oh, Alice},
  journal={arXiv preprint arXiv:2204.04879},
  year={2022}
}

@inproceedings{bibal2022attention,
  title={Is attention explanation? an introduction to the debate},
  author={Bibal, Adrien and Cardon, R{\'e}mi and Alfter, David and Wilkens, Rodrigo and Wang, Xiaoou and Fran{\c{c}}ois, Thomas and Watrin, Patrick},
  booktitle={Proceedings of the 60th Annual Meeting of the Association for Computational Linguistics (volume 1: long papers)},
  pages={3889--3900},
  year={2022}
}

@inproceedings{jain2019attention,
  title={Attention is not explanation},
  author={Jain, Sarthak and Wallace, Byron C},
  booktitle={Proceedings of the 2019 Conference of the North American Chapter of the Association for Computational Linguistics: Human Language Technologies, Volume 1 (Long and Short Papers)},
  pages={3543--3556},
  year={2019}
}

@inproceedings{wiegreffe2019attention,
  title={Attention is not not explanation},
  author={Wiegreffe, Sarah and Pinter, Yuval},
  booktitle={Proceedings of the 2019 conference on empirical methods in natural language processing and the 9th international joint conference on natural language processing (EMNLP-IJCNLP)},
  pages={11--20},
  year={2019}
}

@inproceedings{bai2021attentions,
  title={Why attentions may not be interpretable?},
  author={Bai, Bing and Liang, Jian and Zhang, Guanhua and Li, Hao and Bai, Kun and Wang, Fei},
  booktitle={Proceedings of the 27th ACM SIGKDD conference on knowledge discovery \& data mining},
  pages={25--34},
  year={2021}
}

@article{akula2022attention,
  title={Attention cannot be an explanation},
  author={Akula, Arjun R and Zhu, Song-Chun},
  journal={arXiv preprint arXiv:2201.11194},
  year={2022}
}

@article{geng2026beyond,
  title={Beyond Message Passing: A Symbolic Alternative for Expressive and Interpretable Graph Learning},
  author={Geng, Chuqin and Zhang, Li and Ye, Haolin and Zhao, Ziyu and Jiang, Yuhe and Saba, Tara and Wang, Xinyu and Si, Xujie},
  journal={arXiv preprint arXiv:2602.16947},
  year={2026}
}

@inproceedings{xuanyuan2023global,
  title={Global concept-based interpretability for graph neural networks via neuron analysis},
  author={Xuanyuan, Han and Barbiero, Pietro and Georgiev, Dobrik and Magister, Lucie Charlotte and Li{\`o}, Pietro},
  booktitle={Proceedings of the AAAI conference on artificial intelligence},
  volume={37},
  number={9},
  pages={10675--10683},
  year={2023}
}

@article{farvardin2026end,
  title={End-to-end Feature Alignment: A Simple CNN with Intrinsic Class Attribution},
  author={Farvardin, Parniyan and Chapman, David},
  journal={arXiv preprint arXiv:2603.25798},
  year={2026}
}

@article{yang2025interactive,
  title={Interactive exploration of CNN interpretability via coalitional game theory},
  author={Yang, Lei and Lu, Lingmeng and Liu, Chao and Zhang, Jian and Guo, Kehua and Zhang, Ning and Zhou, Fangfang and Zhao, Ying},
  journal={Scientific Reports},
  volume={15},
  number={1},
  pages={9261},
  year={2025},
  publisher={Nature Publishing Group UK London}
}

@article{mcdonald2009ridge,
  title={Ridge regression},
  author={McDonald, Gary C},
  journal={Wiley Interdisciplinary Reviews: Computational Statistics},
  volume={1},
  number={1},
  pages={93--100},
  year={2009},
  publisher={Wiley Online Library}
}

@article{yang2020revisiting,
  title={Revisiting over-smoothing in deep GCNs},
  author={Yang, Chaoqi and Wang, Ruijie and Yao, Shuochao and Liu, Shengzhong and Abdelzaher, Tarek},
  journal={arXiv preprint arXiv:2003.13663},
  year={2020}
}

@article{rong2025efficient,
  title={Efficient gnn explanation via learning removal-based attribution},
  author={Rong, Yao and Wang, Guanchu and Feng, Qizhang and Liu, Ninghao and Liu, Zirui and Kasneci, Enkelejda and Hu, Xia},
  journal={Acm Transactions on Knowledge Discovery from Data},
  volume={19},
  number={2},
  pages={1--23},
  year={2025},
  publisher={ACM New York, NY}
}

@inproceedings{he2025explaining,
  title={Explaining gnn explanations with edge gradients},
  author={He, Jesse and Rafiey, Akbar and Mishne, Gal and Wang, Yusu},
  booktitle={Proceedings of the 31st ACM SIGKDD Conference on Knowledge Discovery and Data Mining V. 2},
  pages={884--895},
  year={2025}
}

@inproceedings{magister2023concept,
  title={Concept distillation in graph neural networks},
  author={Magister, Lucie Charlotte and Barbiero, Pietro and Kazhdan, Dmitry and Siciliano, Federico and Ciravegna, Gabriele and Silvestri, Fabrizio and Jamnik, Mateja and Li{\`o}, Pietro},
  booktitle={World Conference on Explainable Artificial Intelligence},
  pages={233--255},
  year={2023},
  organization={Springer}
}

\newpage

\appendix

\section{Dataset statistics}
\label{dataset_stats}

Table \ref{dataset_sizes} collates different statistics on the node and graph classification datasets used, such as the number of graphs, average graph size and number of classes.

\begin{table}[h]
\centering
\resizebox{1\textwidth}{!}{%
\begin{tabular}{lccccc}
\hline
\multicolumn{1}{c}{\textbf{Dataset}} &
    \multicolumn{1}{c}{\textbf{\begin{tabular}[c]{@{}c@{}}Classification\\ Problem\end{tabular}}} &
  \multicolumn{1}{c}{\textbf{\begin{tabular}[c]{@{}c@{}}Number of\\ Graphs\end{tabular}}} &
  \multicolumn{1}{c}{\textbf{\begin{tabular}[c]{@{}c@{}}Average\\ Graph Size\end{tabular}}} &
  \multicolumn{1}{c}{\textbf{\begin{tabular}[c]{@{}c@{}}Number of\\ Features\end{tabular}}} &
  \multicolumn{1}{c}{\textbf{\begin{tabular}[c]{@{}c@{}}Number of\\ Classes\end{tabular}}} \\ \hline
\textbf{BA-Shapes}     & Node  & 1    & 700                 & 1  & 4 \\
\textbf{BA-Community}  & Node  & 1    & 1400                & 1  & 8 \\
\textbf{BA-Grid}       & Node  & 1    & 1020                & 1  & 2 \\
\textbf{Tree-Cycles}   & Node  & 1    & 871                 & 1  & 2 \\
\textbf{Tree-Grid}     & Node  & 1    & 1231                & 1  & 2 \\
\hline
\textbf{Grid} & Graph   & 2000   &   22.17  & 1  & 2 \\
\textbf{Grid-House} & Graph  & 1000    &  122.82 & 1  & 2 \\
\textbf{STARS} & Graph       & 1500    & 63.92 & 1  & 3 \\
\textbf{House-Colour} & Graph   & 1000    & 46.95  & 3  & 2 \\
\hline
\textbf{Mutagenicity} & Graph  & 4337 & 30.32  & 14 & 2 \\
\textbf{Reddit-Binary} & Graph & 2000 & 429.63 & 1  & 2 \\ \hline
\end{tabular}%
}
\caption{An overview of key markers of the datasets.}
\label{dataset_sizes}
\end{table}

\section{Model training}
\label{training}

Table \ref{training} summarizes the model hyperparameters, following the settings proposed by \citet{magister2023concept} and \citet{longa2022explaining}. We train all models using stochastic gradient descent \citep{pmlr-v70-li17f} across five different seeds to allow reporting confidence intervals. Our implementation is in PyTorch \citep{paszke2019pytorch}.

\begin{table*}[h]
\centering
\resizebox{\textwidth}{!}{
\begin{tabular}{lcccccccc}
\hline
\textbf{Dataset} &
  \textbf{\begin{tabular}[c]{@{}c@{}}Number of \\ Graph Conv.\end{tabular}} &
  \textbf{\begin{tabular}[c]{@{}c@{}}Number of \\ Hidden Units\end{tabular}} &
  \textbf{\begin{tabular}[c]{@{}c@{}}Node Concept \\ Encoding Size\end{tabular}} &
  \textbf{\begin{tabular}[c]{@{}c@{}}Learning \\ Rate\end{tabular}} &
  \textbf{\begin{tabular}[c]{@{}c@{}}Batch \\ Size\end{tabular}} &
  \textbf{\begin{tabular}[c]{@{}c@{}}Number of \\ Epochs\end{tabular}} \\ \hline
\textbf{BA-Shapes}     & 4 & 10 & 10 & 0.001  & - & 7000  \\ 
\textbf{BA-Grid}       & 4 & 10 & 10 & 0.001  & - & 3000  \\ 
\textbf{Tree-Grid}     & 7 & 20 & 20 & 0.0001 & - & 20000 \\ 
\textbf{Tree-Cycle}    & 3 & 10 & 10 & 0.001  & - & 7000  \\ 
\textbf{BA-Community}  & 6 & 20 & 20 & 0.001 & - & 10000 \\
\hline
\textbf{Grid}    & 5   &   20  & 10  &  0.001 & 16 & 300 \\
\textbf{Grid-House}   & 4    &  20 & 10  &  0.001 & 16 & 2000 \\
\textbf{STARS}        & 2    & 10 & 4  &  0.001 & 16 & 300 \\
\textbf{House-Colour}    & 2    & 10  & 10 & 0.001 & 16 & 300 \\
\textbf{Mutagenicity}   & 3 & 40 & 10 &  0.001 & 16 & 1000 \\
\textbf{Reddit-Binary}  & 3 & 32 & 10  &  0.001 & 16 & 1000 \\ \hline
\end{tabular}%
}
\caption{A summary of the model architecture and training hyperparameters for each dataset, following recommendations of \citet{magister2023concept} and \citet{longa2022explaining}. These parameters are kept the same between all models trained.}
\label{training}
\end{table*}

\section{Concept evolution in graph classification}
\label{graph_class_evolution}

\begin{figure}[h!]
    \centering
    \subfloat[Concept Completeness Score over \textbf{CGC} Layers for Node Classification]{
        \includegraphics[width=0.47\textwidth, trim=0.25cm 0.4cm 0.25cm 1.5cm, clip]{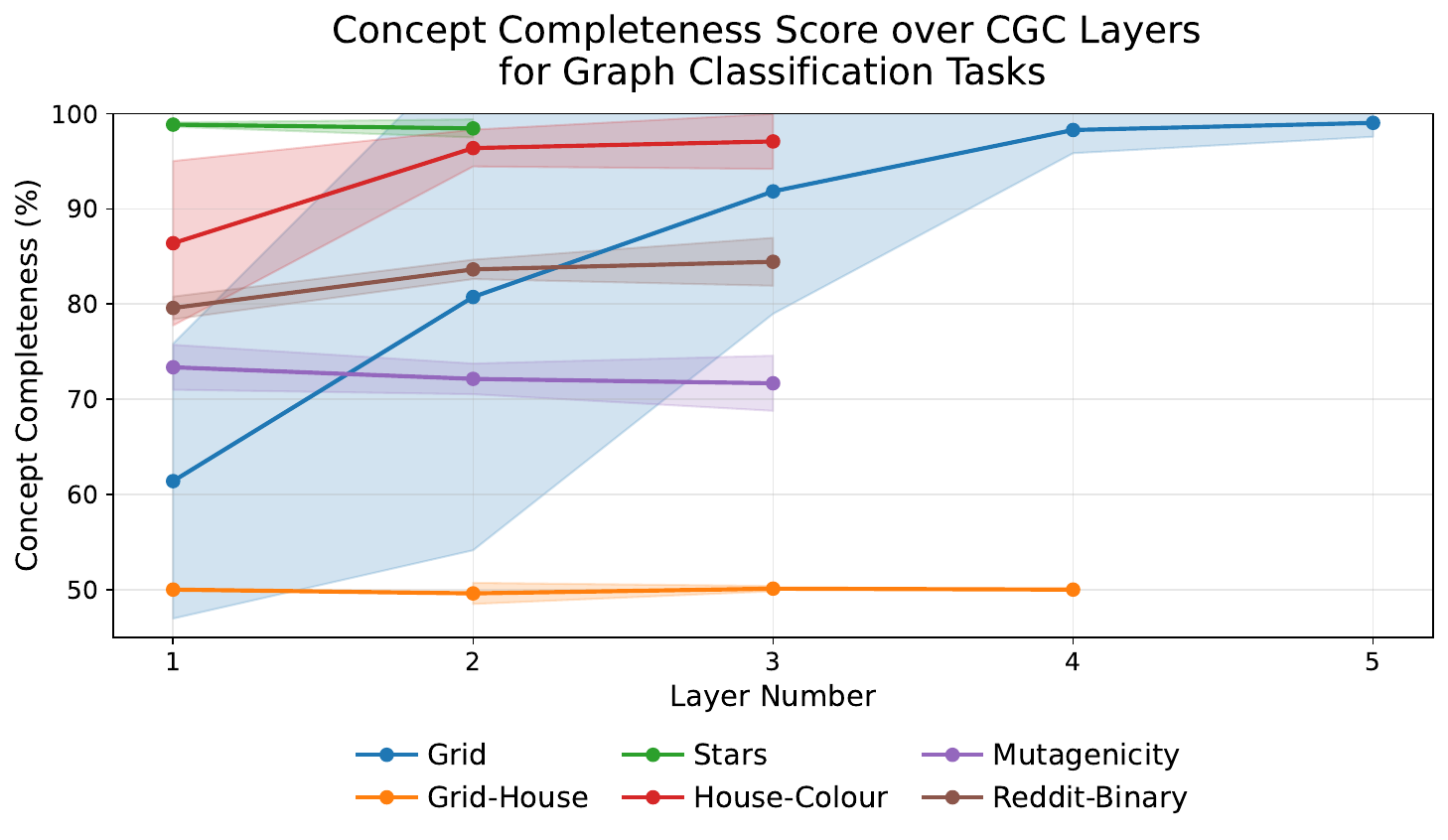}
    }
    \hfill
    \subfloat[Concept Completeness Score over \textbf{Pure CGC} Layers for Node Classification]{
        \includegraphics[width=0.47\textwidth, trim=0.25cm 0.4cm 0.25cm 1.5cm, clip]{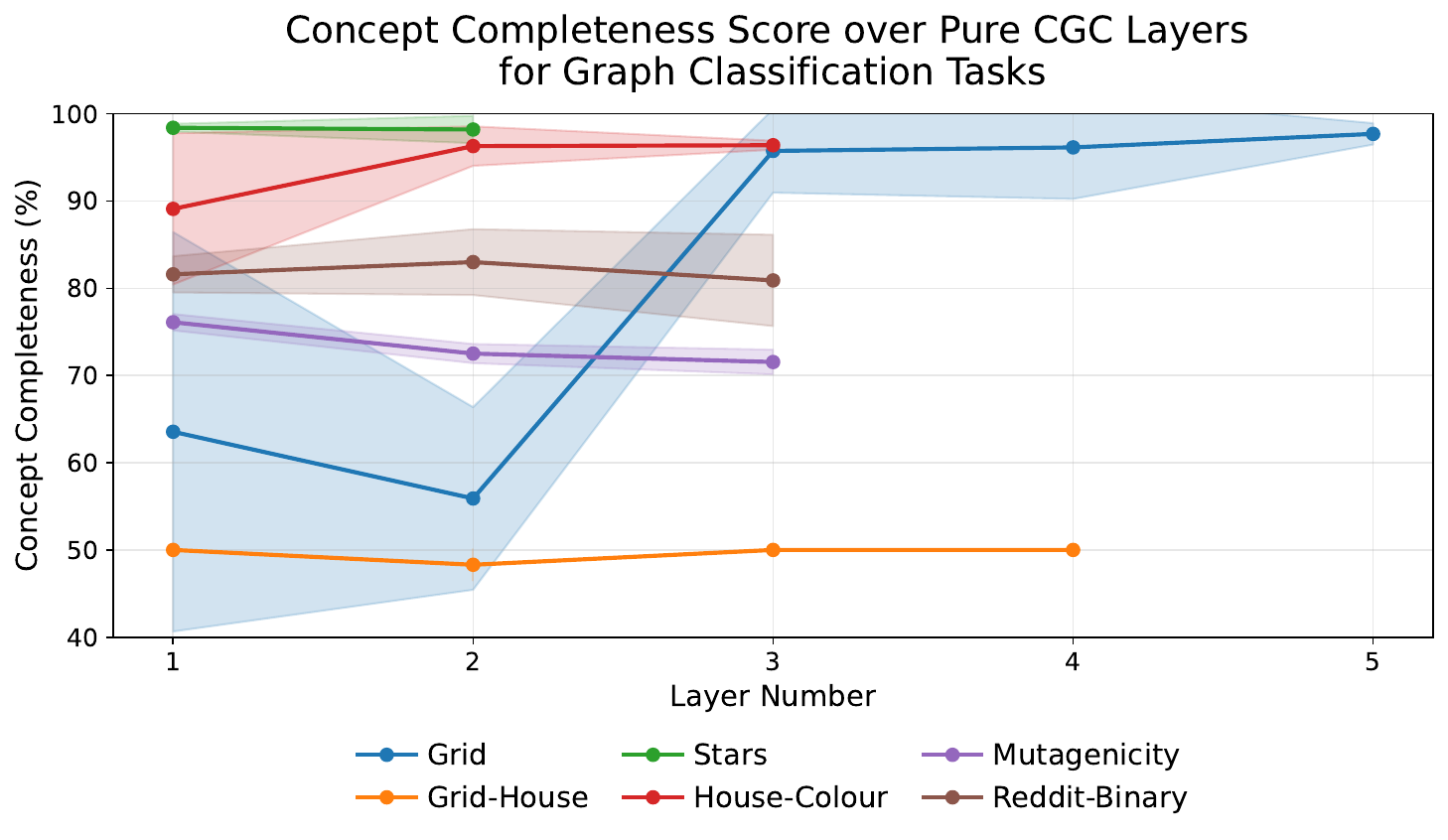}
        \label{concept_completeness_pure_layer}
    }
    \caption{\raggedright Concept completeness measured across Concept Graph Convolution (CGC) and pure CGC layers trained on different graph classification tasks. We can observe that concept completeness increases with model depth, indicating that the concepts representing a greater receptive field more accurately represent the task. On the Reddit-Binary and Mutagenicity dataset, we can observe a slight decline of concept completeness with model depth.}
    \label{fig:concept_completeness_plot_graph}
\end{figure}

We also observe an increase in the concept completeness scores across CGC layers in graph classification tasks, as shown in Figure \ref{fig:concept_completeness_plot_graph}. We can see for simpler tasks like the STARS dataset, concept completeness is very high from the start, validating the shallow network design. In contrast, we do not observe any significant improvement in concept completeness with model depth for the Mutagenicity and Reddit-Binary datasets. This could potentially indicate that the correct number of layers has been used or that a wider network may be needed to represent a more nuanced concept space to increase concept completeness and model accuracy.

\section{Effects of oversmoothing on the concept space}
\label{oversmoothing}

\begin{figure}[h!]
    \centering
    \subfloat[Concept Completeness across an increasing number of CGC layers]{
        \includegraphics[width=0.49\textwidth, trim=0.27cm 0.4cm 0.31cm 1cm, clip]{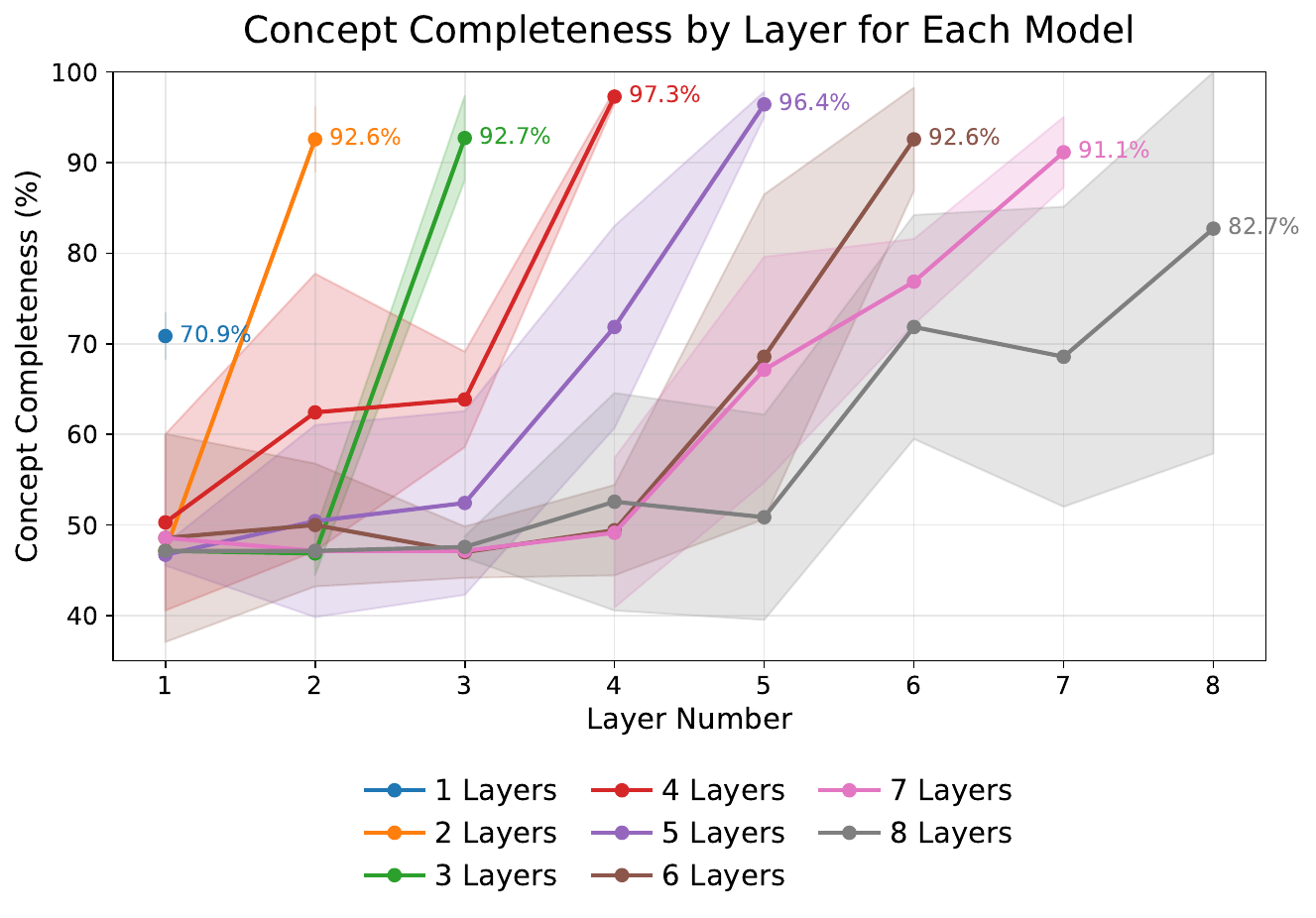}
    }
    \hfill
    \subfloat[Model Accuracy for models with an increasing number of CGC layers]{\raisebox{6mm}{
        \includegraphics[width=0.47\textwidth, trim=0.25cm 0cm 0.25cm 1cm, clip]{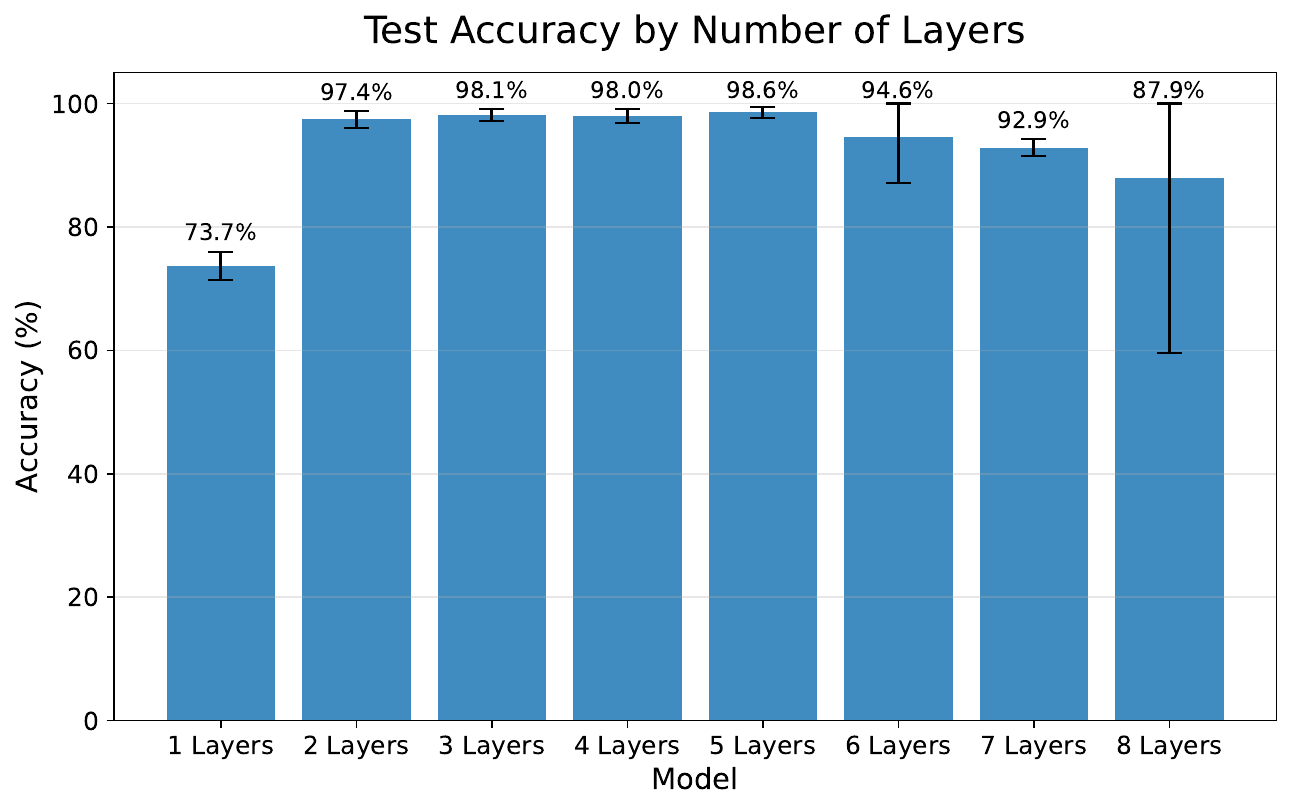}}
    }
    \caption{\raggedright We measure concept completeness across models with an increasing number of Concept Graph Convolution (CGC) layers, trained on the BA-Shapes node classification dataset.}
    \label{fig:deeper_layers_fig}
\end{figure}

We observe that concepts distilled from later convolutional layers are more representative of the final task. This leads us to question how oversmoothing affects the concept space. It can be hypothesized that concept completeness declines with oversmoothing for two reasons: (i) task accuracy declines and concept completeness is bounded by it \citep{magister2021gcexplainer} and (i) representations collapse as the receptive field grows. To explore this, we train models with an increasing number of layers for the BA-Shapes dataset and observe the behaviour of concept completeness across layers. Figure \ref{fig:deeper_layers_fig} shows concept completeness across model layers, as well task accuracy for models with a varying number of layers. We find that task accuracy declines for models with more than 5 layers, which is mirrored by the deterioration of the concept completeness score. Moreover, we observe that concept completeness at early layers in these models generally is lower than in more shallow networks, which can be attributed to shallow layers receiving less of an update during gradient descent. Most notably, we also observe that concept completeness is highest for the model with three layers, while accuracy peaks with the model using 5 layers. This indicates that more shallow networks may provide the best trade-off between high accuracy and strong concept completeness. This is further underlined by the confidence interval on the concept completeness score for the model with only 3 layers being much lower in the final layers than the other models. While representations may not be oversmoothed, the set of concepts discovered by the shallower model is more complete, providing an improved insight at only a loss of 0.5\% accuracy points.

\section{Concept evolution over training time}
\label{training_time}

\begin{figure}[h!]
    \centering
    \subfloat[Concept completeness of the \textbf{CGC} layers across epochs]{
        \includegraphics[width=0.45\textwidth, trim=0.2cm 0.4cm 0.2cm 0.7cm, clip]{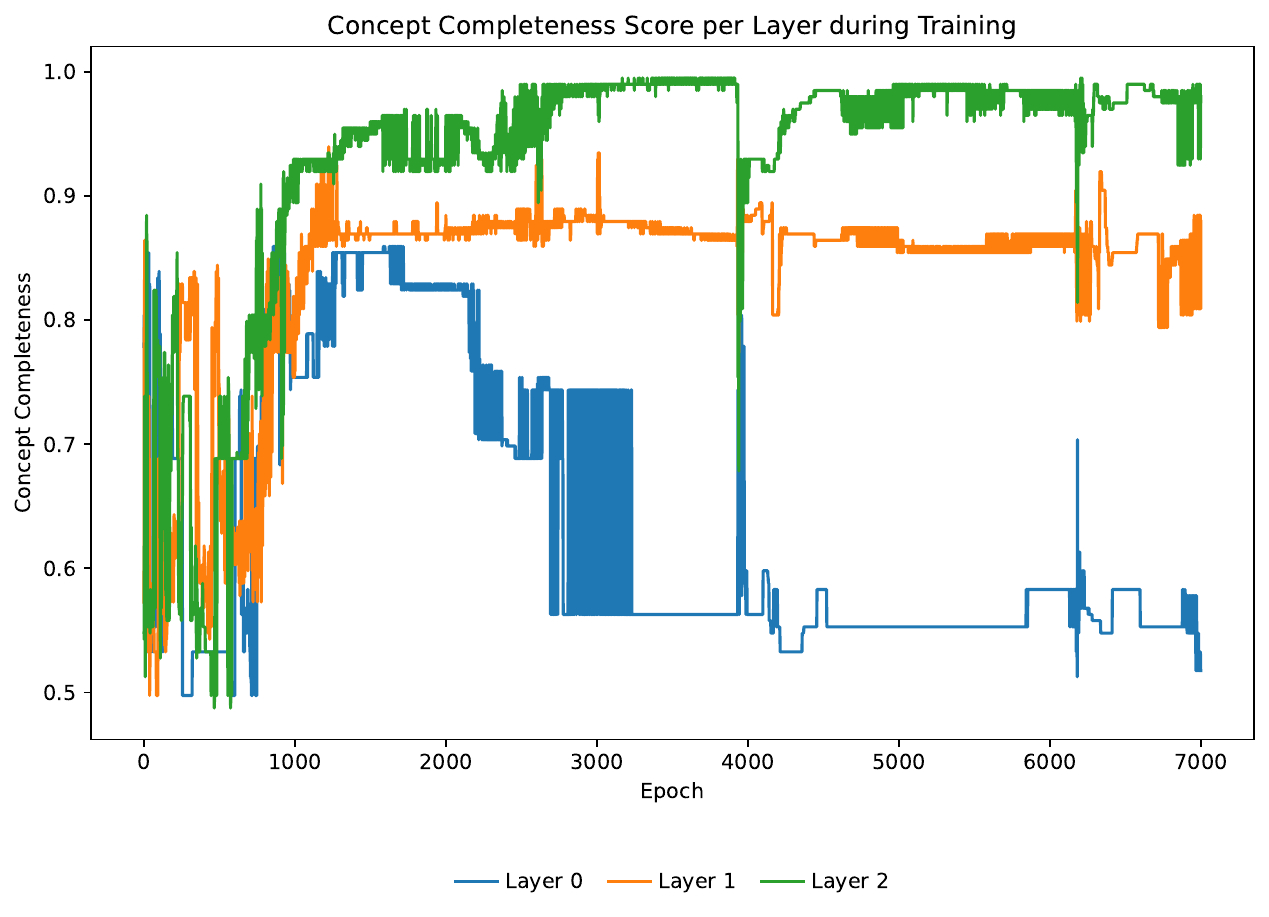} \label{training_a}}
    \hfill
    \subfloat[\textbf{CGC} model accuracy across epochs]{\raisebox{5mm}{
        \includegraphics[width=0.52\textwidth, trim=0.5cm 0.5cm 0.5cm 1.7cm, clip]{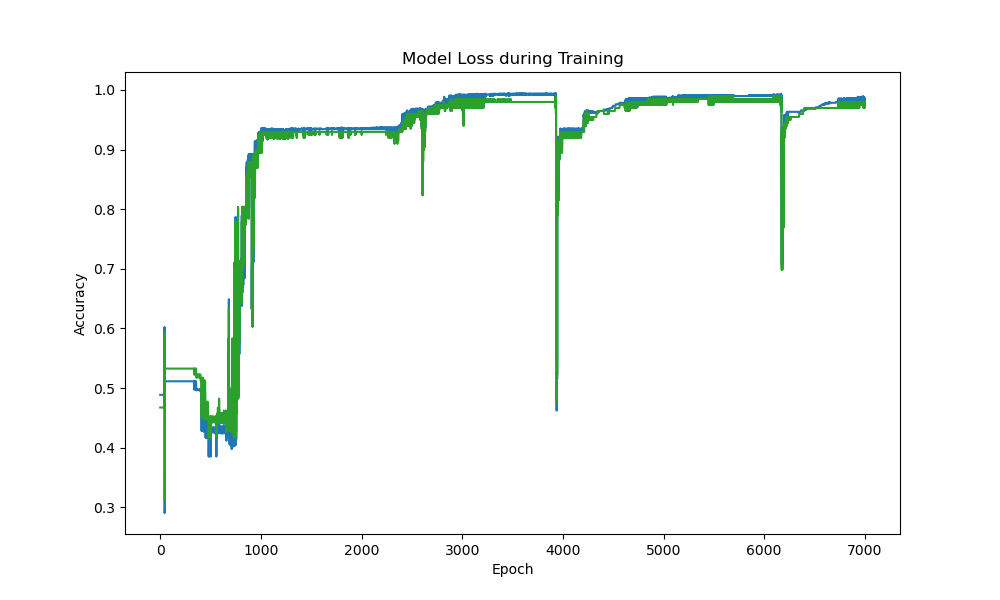}}\label{training_b}} 
    \\
    \centering
    \subfloat[Concept completeness of the \textbf{pure CGC} layers across epochs]{
        \includegraphics[width=0.45\textwidth, trim=0.2cm 0.4cm 0.2cm 0.7cm, clip]{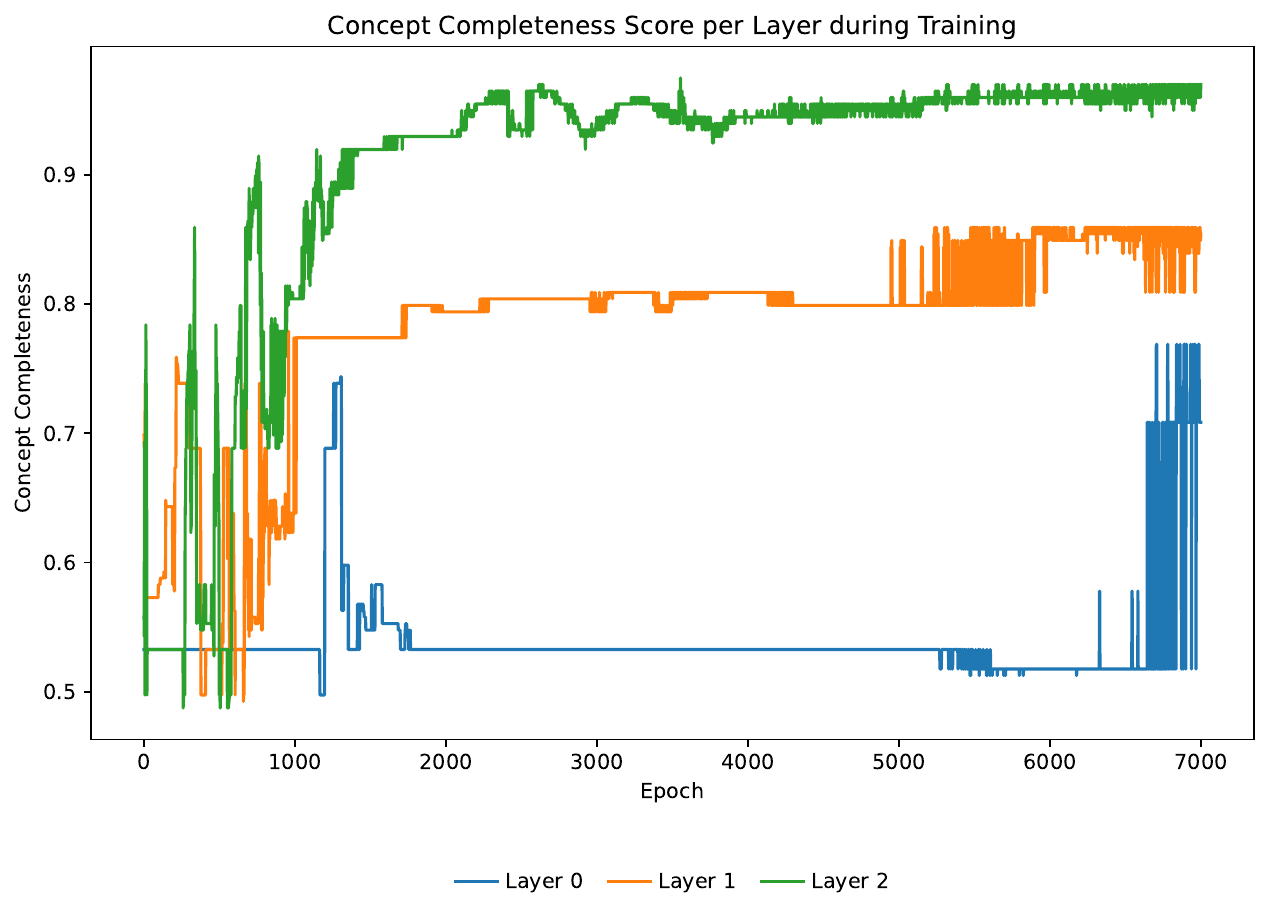}
        \label{training_c}}
    \hfill
    \subfloat[\textbf{Pure CGC} model accuracy across epochs]{\raisebox{5mm}{
        \includegraphics[width=0.52\textwidth, trim=0.5cm 0.5cm 0.5cm 1.7cm, clip]{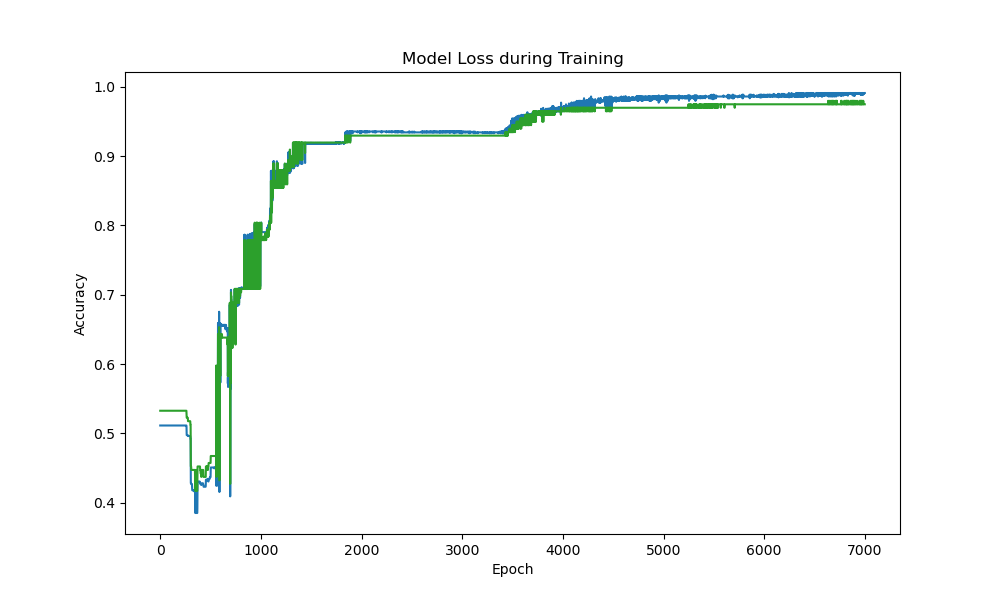}}\label{fig:training_d}}
    \caption{\raggedright Concept completeness and accuracy measured across training epochs for a model using Concept Graph Convolution (CGC) and pure CGC layers for classifying the Tree-Cycle dataset. We observe fluctuations in concept completeness during training, related to fluctuations in accuracy.}
    \label{fig:training_concept_completeness_tree_cycle}
\end{figure}

We also observe the concept completeness score over training epochs. We are interested to confirm the observation by \citet{xuanyuan2023global}, which finds that the interpretability of a model decreases after a certain point in model training. Figure \ref{fig:training_concept_completeness_tree_cycle} plots the concept completeness score of different layers across epochs, as well as the accuracy of a model trained on the Tree-Cycles dataset with CGC and pure CGC layers, respectively. In Figure \ref{training_a}, we observe that the concept completeness of the CGC layers increases for $\sim1500$ epochs, after which we observe a collapse in concept completeness in Layer 0, while concept completeness for other layers remains stable and accuracy increases (Figure \ref{training_b}). While the concept completeness of the remaining layers remains high, we can see this collapse in the concept completeness score of Layer 0 as interpretability declining with more epochs. We would therefore recommend selecting a checkpoint before epoch 1500, as both model accuracy and interpretability are high. In general, we observe that concept completeness at later layers and model accuracy are tied. The sharp drop in accuracy at epoch $\sim4000$ is mirrored in a decline in concept accuracy in the last layer. However, concept completeness in the first layer appears to jump up indicating that the model may have reached an suboptimal concept representation state for layer 0. Figures \ref{training_c} and \ref{fig:training_d}, depict concept completeness and accuracy across epochs for a model using pure CGC layers. Overall, the training of the model using pure CGC layers appears more stable. Nevertheless, concept completeness for Layer 0 also remains low here.

\captionsetup[subfloat]{justification=centering}

\section{Behaviour of $\gamma$ in models trained for graph classification}
\label{gamma}

\begin{figure}[h!]
    \centering
    \subfloat[Gamma across layers for \textbf{CGC} layers in graph classification models]{
        \includegraphics[width=0.48\textwidth, trim=0.25cm 0.4cm 0.25cm 1.8cm, clip]{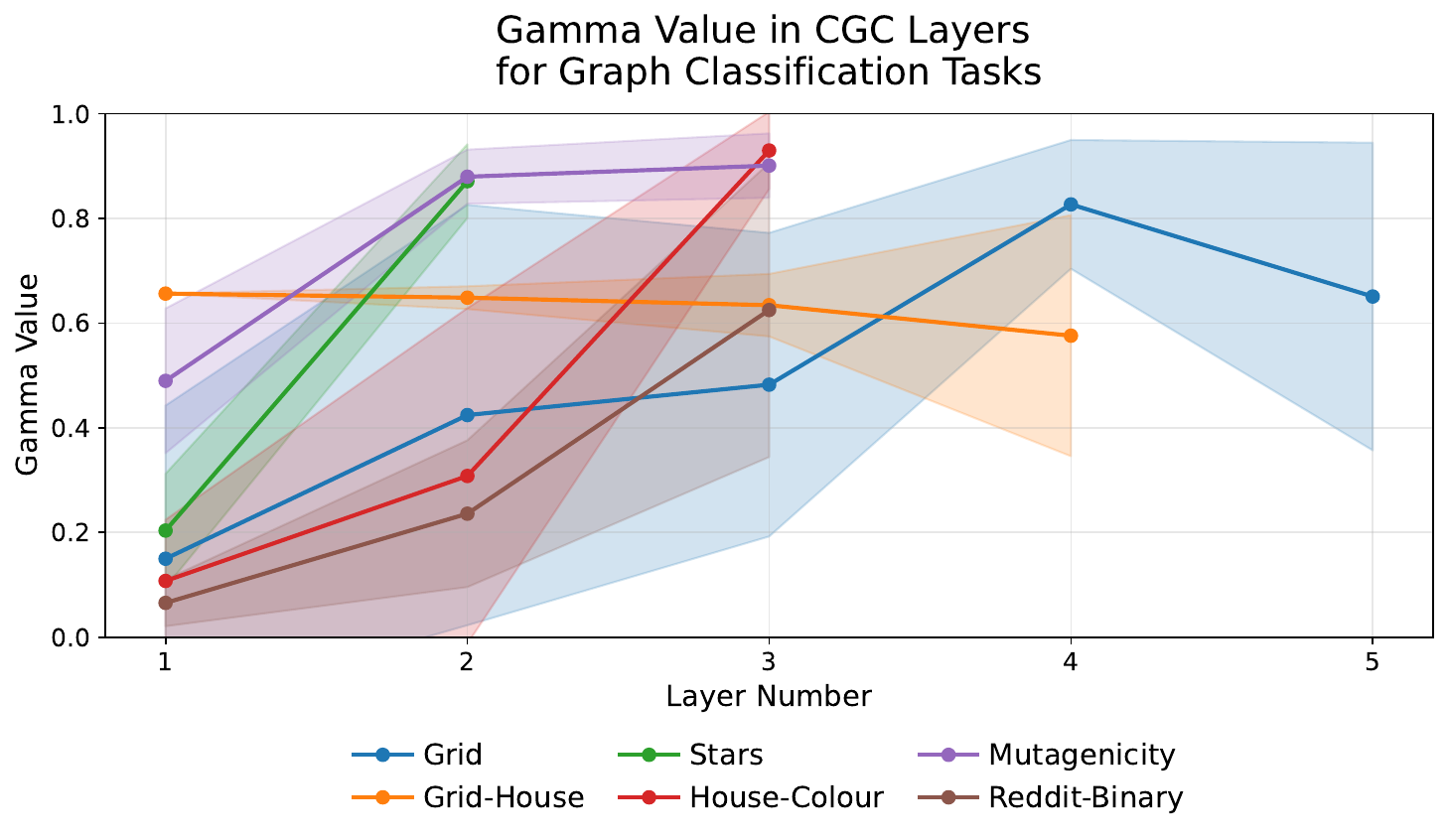}
        \label{fig:gamma_c}
    }
    \hfill
    \subfloat[Gamma across layers for \textbf{pure CGC} layers in graph classification models]{
        \includegraphics[width=0.48\textwidth, trim=0.25cm 0.4cm 0.25cm 1.8cm, clip]{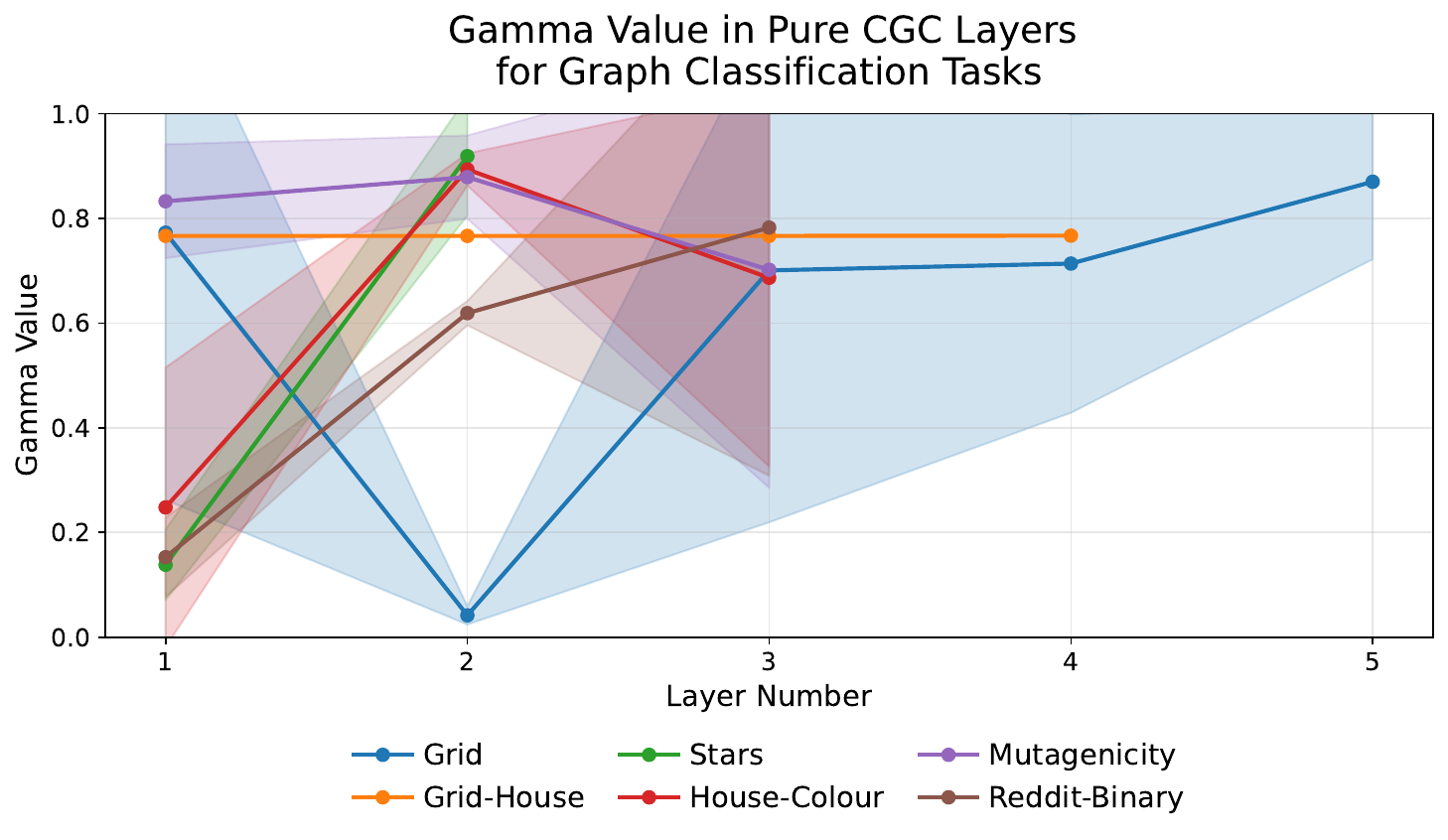}
        \label{fig:gamma_d}
    }
    \caption{\raggedright We plot the learned $\gamma$ value of Concept Graph Convolution (CGC) and pure CGC layers across the models trained for graph classification.}
    \label{fig:gamma2}
\end{figure}

In Figure \ref{fig:gamma2}, we see a stronger upwards trend in the value of $\gamma$ across CGC layers in models for graph classification. For example, Figure \ref{fig:gamma_c} plots the $\gamma$ value of CGC layers across models trained on different graph classification tasks. Across the datasets, we can observe an upwards trend in the $\gamma$ value with all models having $\gamma$ value greater than $0.5$ in the final layer. This means that the attention on only a few neighbouring concepts becomes increasingly important as the receptive field grows. Here, simple structural aggregation may drown out the signal of important concepts, as we use sparse, fuzzy concept representations. We do not observe a trend as strong for the pure CGC layer plotted in Figure \ref{fig:gamma_d}, as the $\gamma$ value fluctuates more between layers. Moreover, for the Mutagenicity and House-Colour dataset, we observe a decline in the $\gamma$ value in the last layer in line with our observations for the pure GCG layer when used for node classification. Overall, our analysis supports that a learned $\gamma$ parameter allows the model the required flexibility for aggregating sparse concept representations.

\section{Reliance on node embeddings versus concept embeddings}
\label{eta}

\begin{figure}[h!]
    \centering
    \subfloat[Eta across layers for \textbf{CGC} layers in node classification models]{
        \includegraphics[width=0.48\textwidth, trim=0.25cm 0.4cm 0.25cm 1.8cm, clip]{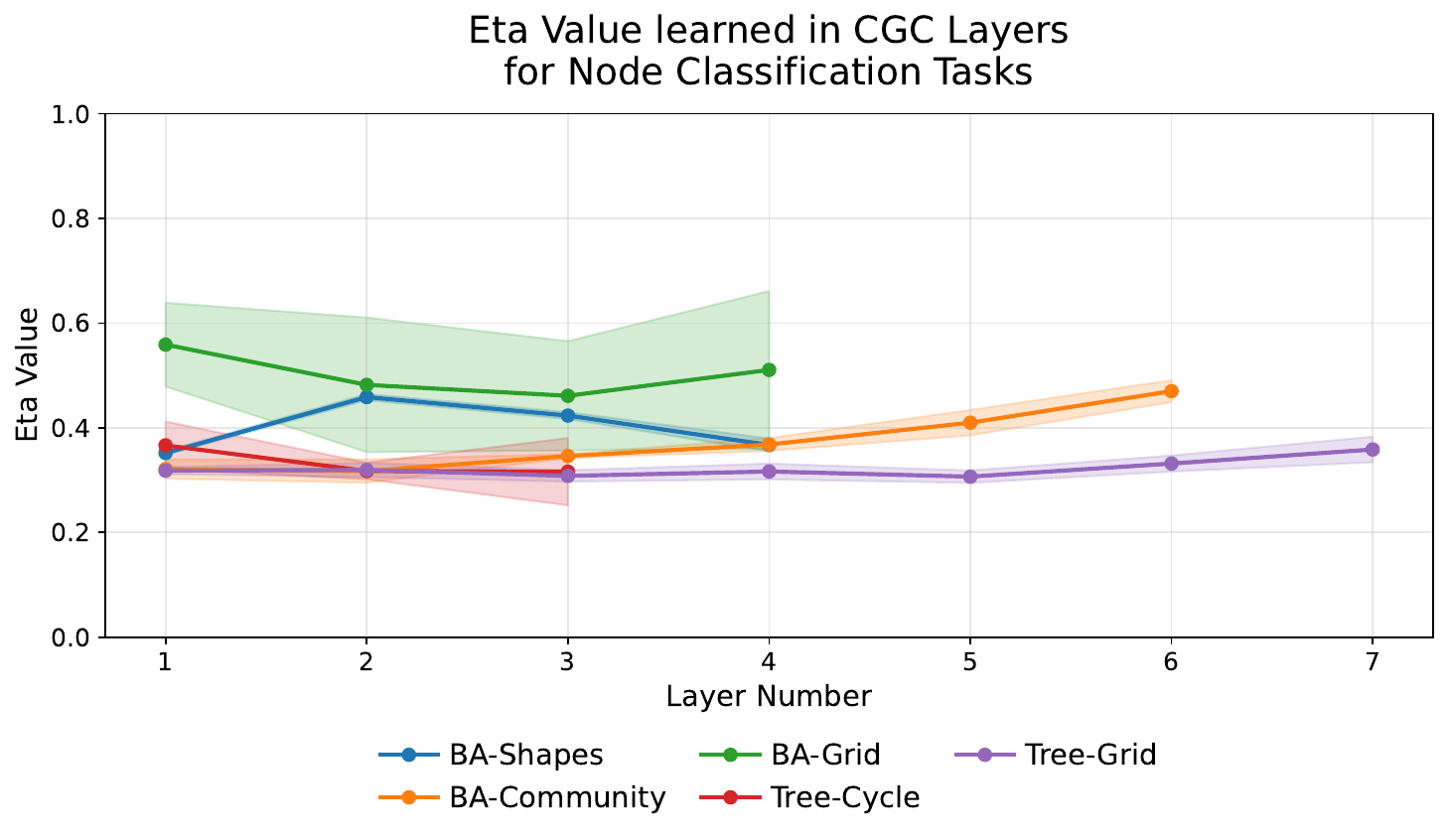}
        \label{fig:eta_a}
    }
    \hfill
    \subfloat[Eta across layers for \textbf{CGC} layers in graph classification models]{
        \includegraphics[width=0.48\textwidth, trim=0.25cm 0.4cm 0.25cm 1.7cm, clip]{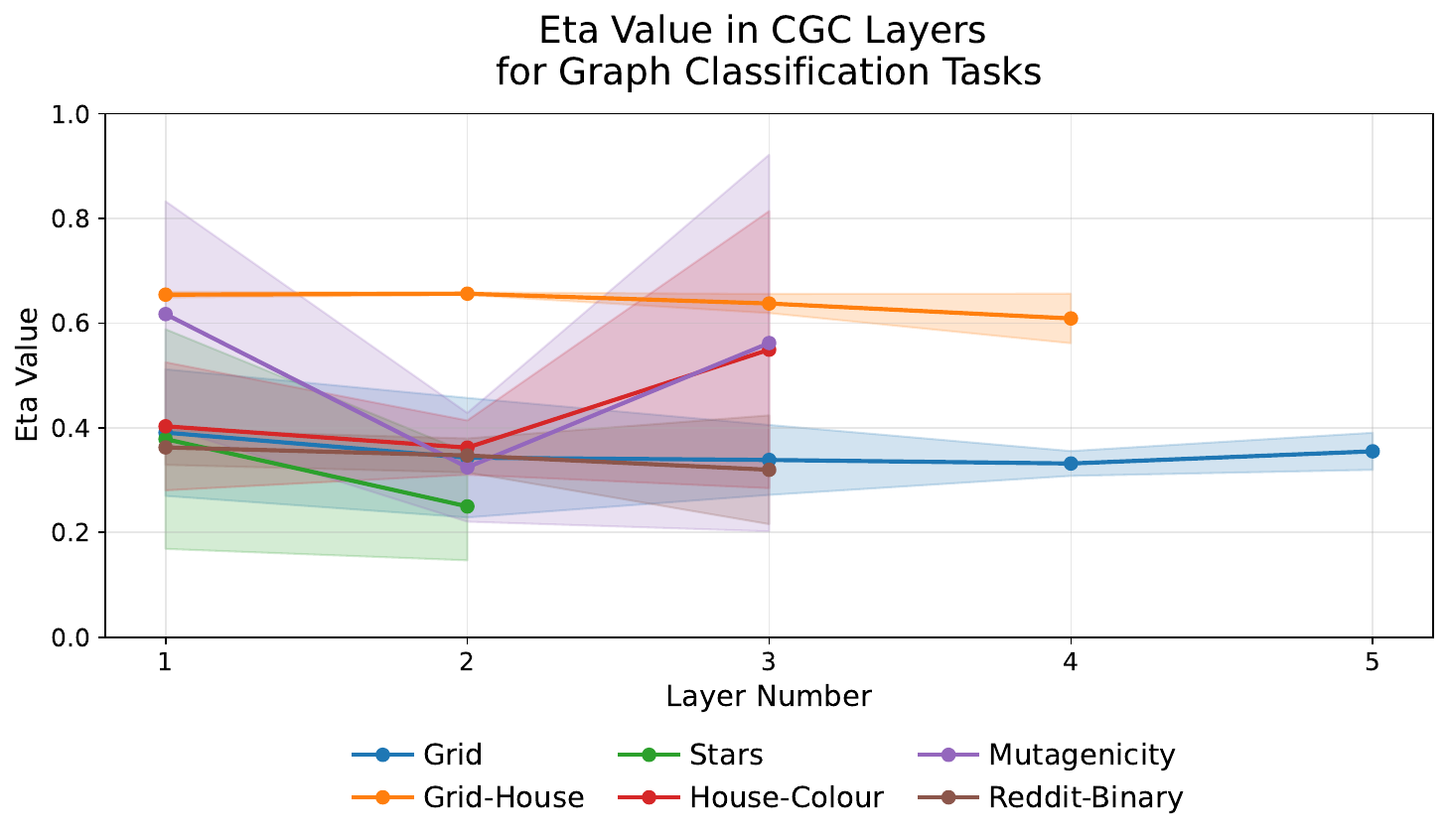}
        \label{fig:eta_b}
    }
    \caption{\raggedright We plot the learned $\eta$ value of Concept Graph Convolution (CGC) and pure CGC layers across the models trained for node and graph classification.}
    \label{fig:eta}
\end{figure}

While the pure CGC layer only operates in the node concept space, the CGC layer also learns a mixing parameter $\eta$ for determining the mixing of the raw node representations with the concept node representations. Figure \ref{fig:eta} plots the value of $\eta$ across model layers in models for node and graph classification, respectively. We find that eta remains fairly stable in node classification, with four out of five tasks having $\eta < 0.5$, focusing more on the raw node representations. However, $\eta$ does not fall below $\sim0.25$ in node classification, indicating that the model relies both on the unconstrained and concept latent space. We observe a similar trend across the models trained for graph classification. Only the Grid-House dataset appears to have a higher $\eta$ value, however, the model does not learn to predict the task well and this $\eta$ value my only indicate that the model learns to increase $\eta$ to decrease the loss, as we regularize for a higher $\eta$ for improved interpretability. Overall, it can be argued that the pure CGC layer is more interpretable as the network's computation is not obscured by raw latent representations as in the GCG layer, which appears to be influenced slightly more by raw latent representations than the concept representations.

\section{Concept visualisations}
\label{vis}

\begin{figure}[h]
\centerline{\includegraphics[width=1\textwidth]{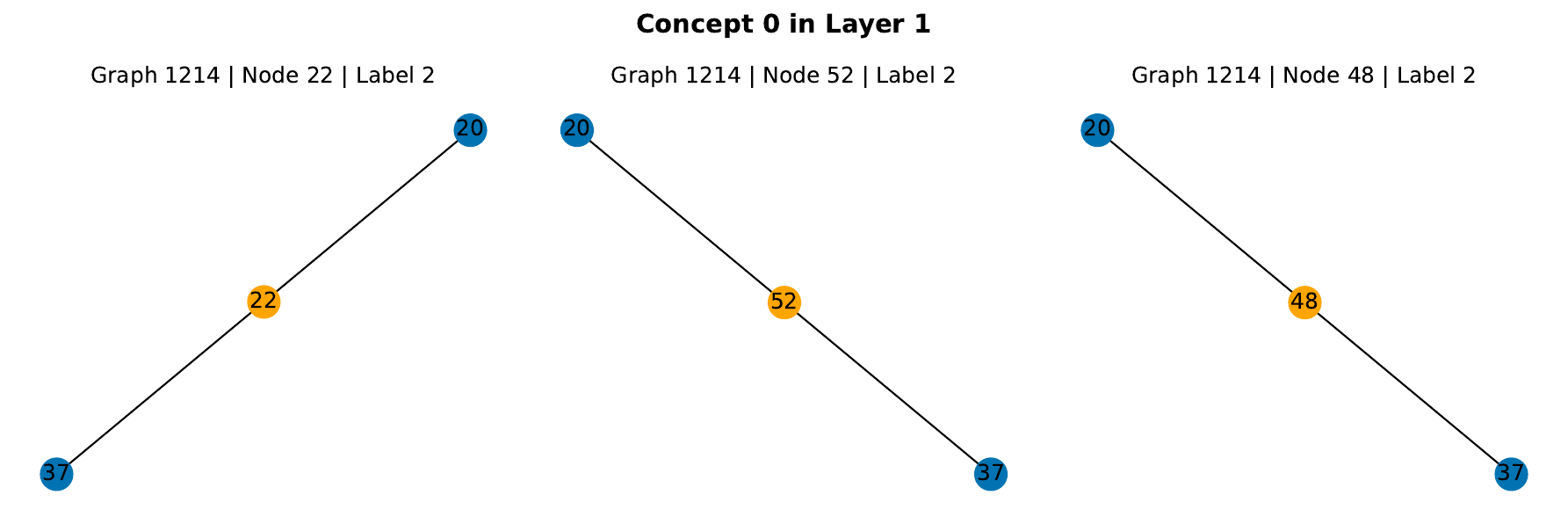}}
\caption{A concept discovered in the first layer of a model using pure Concept Graph Convolution (CGC) layers, trained on the STARS dataset. The orange nodes represent the nodes belonging to the concept, while the blue nodes are the expanded neighbourhood.}
\label{cgc_stars1}
\end{figure}

\begin{figure}[h]
\centerline{\includegraphics[width=1\textwidth]{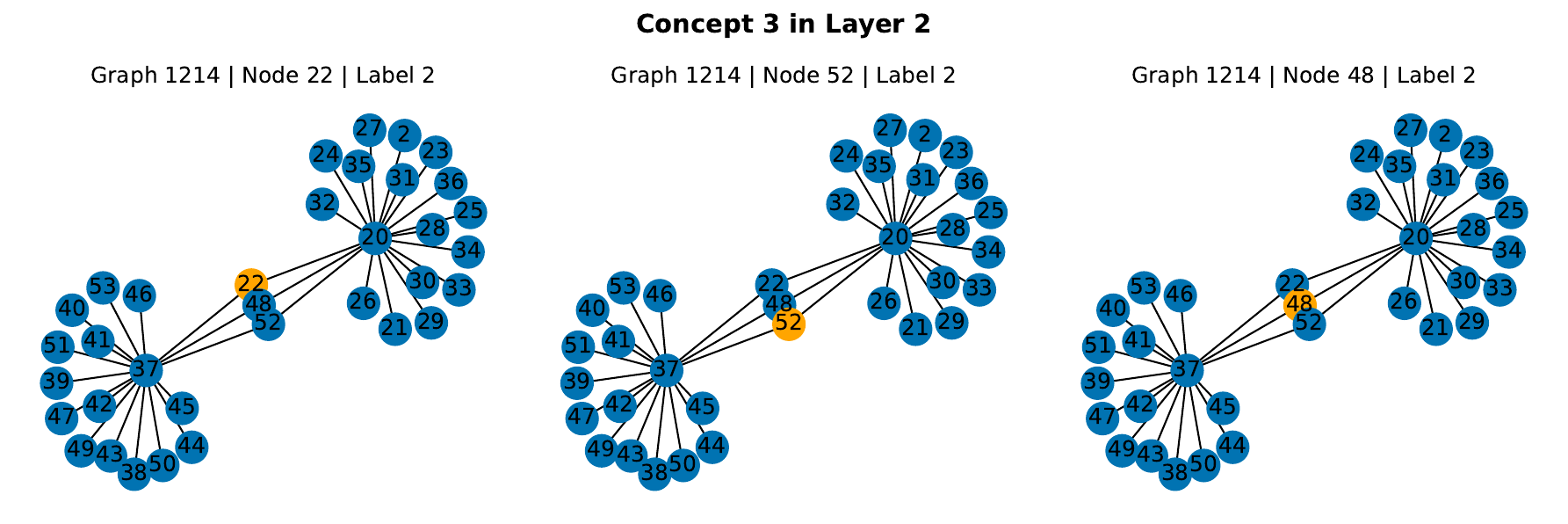}}
\caption{A concept discovered in the second layer of a model using pure Concept Graph Convolution (CGC) layers, trained on the STARS dataset. The orange nodes represent the nodes belonging to the concept, while the blue nodes are the expanded neighbourhood.}
\label{cgc_stars2}
\end{figure}

The CGC layers allow us to visualize concepts discovered at different stages of the receptive fields of a GNN, as we can extract concepts per layer. A benefit of extracting concepts across layers is that we can examine the evolution of concepts, specifically how a concept morphs into a larger concept, as the receptive field grows. We visualize concepts as proposed by \citet{magister2021gcexplainer} via the $p$-hop neighbourhood of the clustered nodes, where $p$ is the appropriate layer depth. For example, Figure \ref{cgc_stars1} shows an early concept discovered for the STARS dataset, showing a node connected to to exactly two other nodes. When we examine Figure \ref{cgc_stars2} in relation to it, we can see that we have now discovered the larger concept for these nodes. Notice, how both concepts highlight the same nodes in Graph 1214, which have a mirrored structure. While the concept extracted at the first layer only showed that these nodes have exactly two neighbours, the concept extracted at the second layer shows that these nodes all attach to the same two star structures. As all three visualizations come from the same graph, we can identify this as a rare, fine-grained concept. Note, that the concept label between layers changes, as we extract it from the binarized vector.

\begin{figure}[h]
\centerline{\includegraphics[width=1\textwidth]{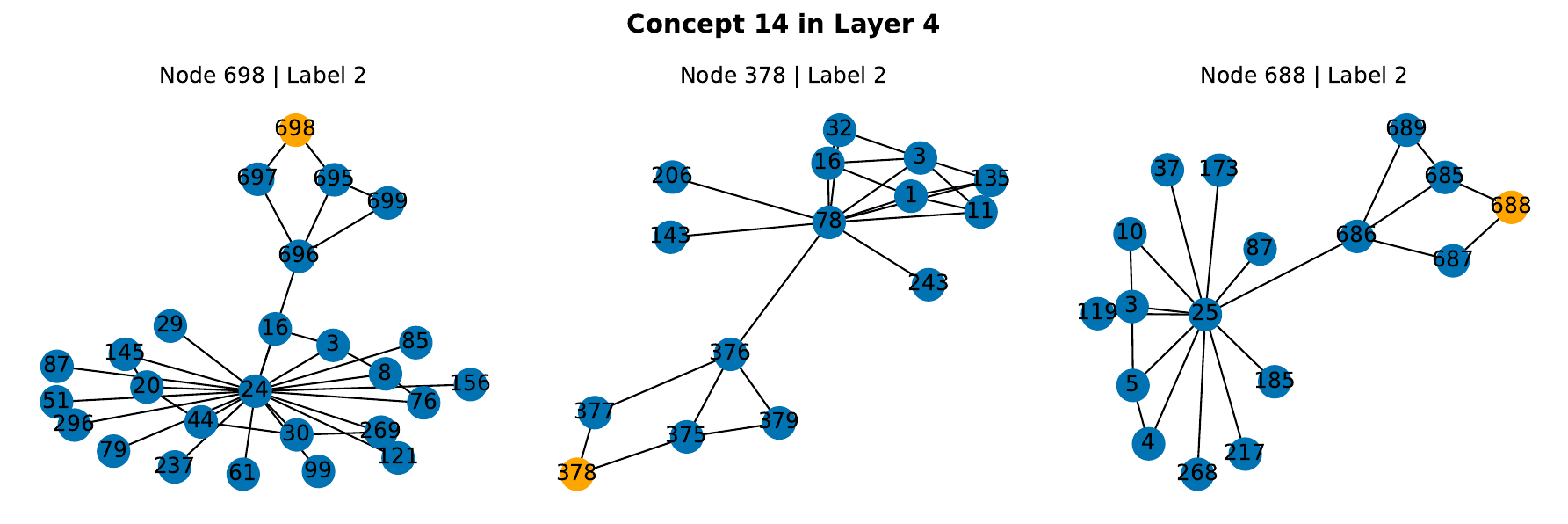}}
\caption{A concept discovered in the fourth layer of a model using Concept Graph Convolution (CGC) layers, trained on the BA-Shapes dataset. The orange nodes represent the nodes belonging to the concept, while the blue nodes are the expanded neighbourhood.}
\label{cgc_ba_shapes1}
\end{figure}

\begin{figure}[h]
\centerline{\includegraphics[width=1\textwidth]{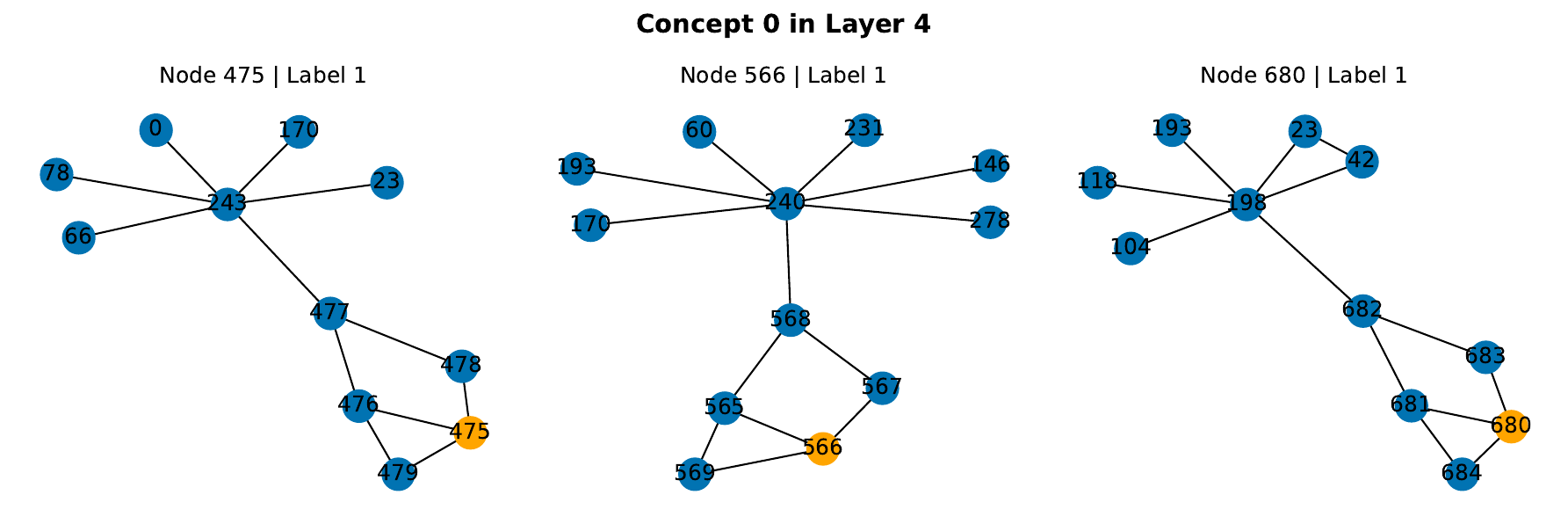}}
\caption{A concept discovered in the fourth layer of a model using \textbf{pure} Concept Graph Convolution (CGC) layers, trained on the BA-Shapes dataset. The orange nodes represent the nodes belonging to the concept, while the blue nodes are the expanded neighbourhood.}
\label{cgc_ba_shapes2}
\end{figure}

\begin{figure}[h]
\centerline{\includegraphics[width=0.3\textwidth]{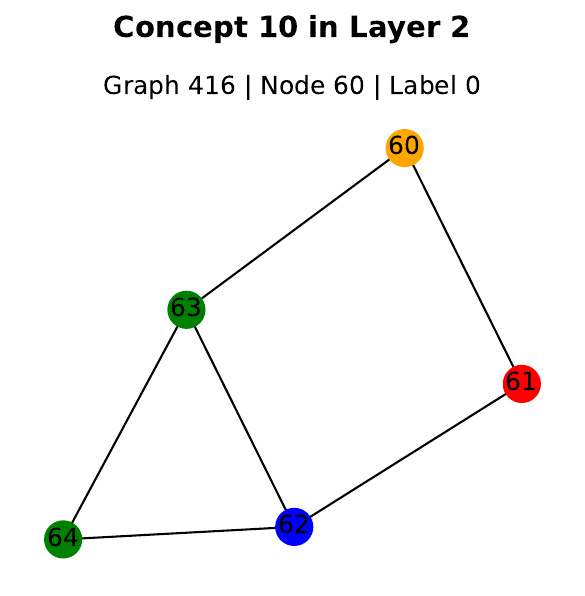}}
\caption{A rare concept discovered in the second layer of a model using \textbf{pure} Concept Graph Convolution (CGC) layers, trained on the House-Colour dataset. The orange node represent the node belonging to the concept, while the other nodes are coloured in accordance with the dataset.}
\label{cgc_house_colour2}
\end{figure}

We also successfully learn fine-grained and rare concepts across layers in models for other datasets. For example, Figure \ref{cgc_ba_shapes1} shows that we can distil fine-grained concepts with the CGC, like the bottom node of a house structure at the far side of the middle node attached to the base structure for the BA-Shapes dataset. We observe that we can also identify fine-grained concepts with the pure CGC implementation, as shown in Figure \ref{cgc_ba_shapes2}, depicting a concept representing the middle node of the house opposite to the bottom node attached to the base graph in the BA-Shapes dataset. Lastly, we can also identify rare concepts, which are concepts with only a small number of nodes assigned. For example, Figure \ref{cgc_house_colour2} shows a singular house structure with unique random colouring in the House-Colour dataset.

\begin{figure}[h]
\centerline{\includegraphics[width=1\textwidth]{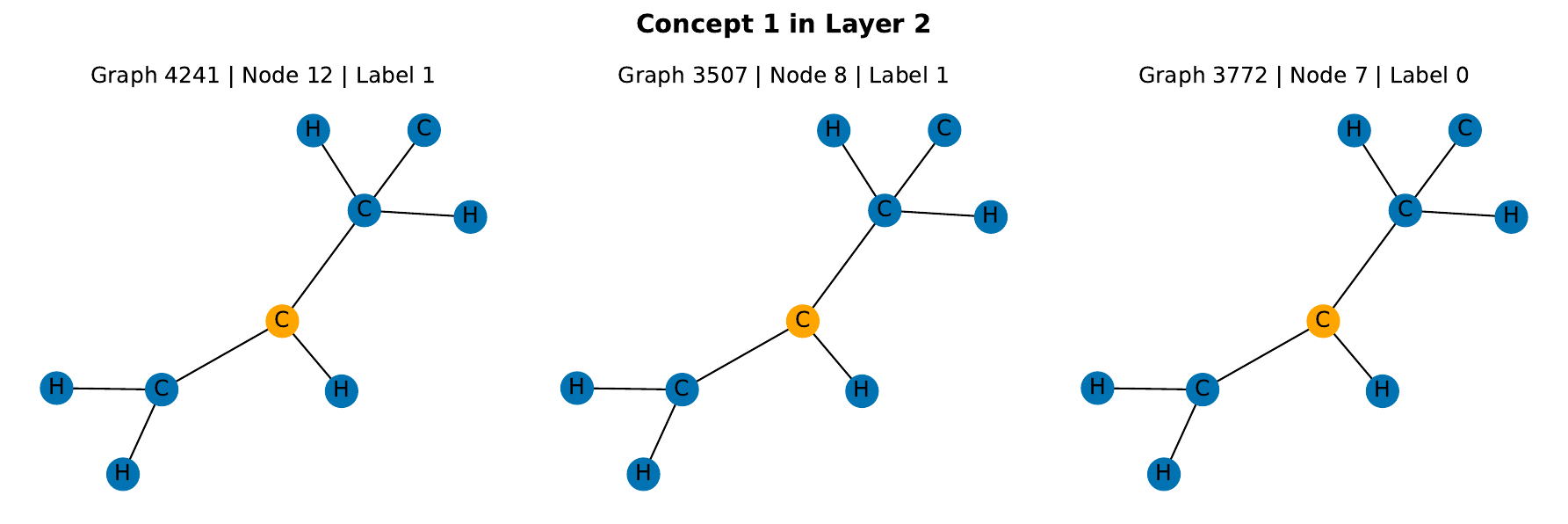}}
\caption{A concept discovered in the second layer of a model using Concept Graph Convolution (CGC) layers, trained on the Mutagenicity dataset. The orange nodes represent the nodes belonging to the concept, while the blue nodes are the expanded neighbourhood. Node labels correspond to the atom label.}
\label{cgc_mutag_concept1}
\end{figure}

\begin{figure}[h]
\centerline{\includegraphics[width=1\textwidth]{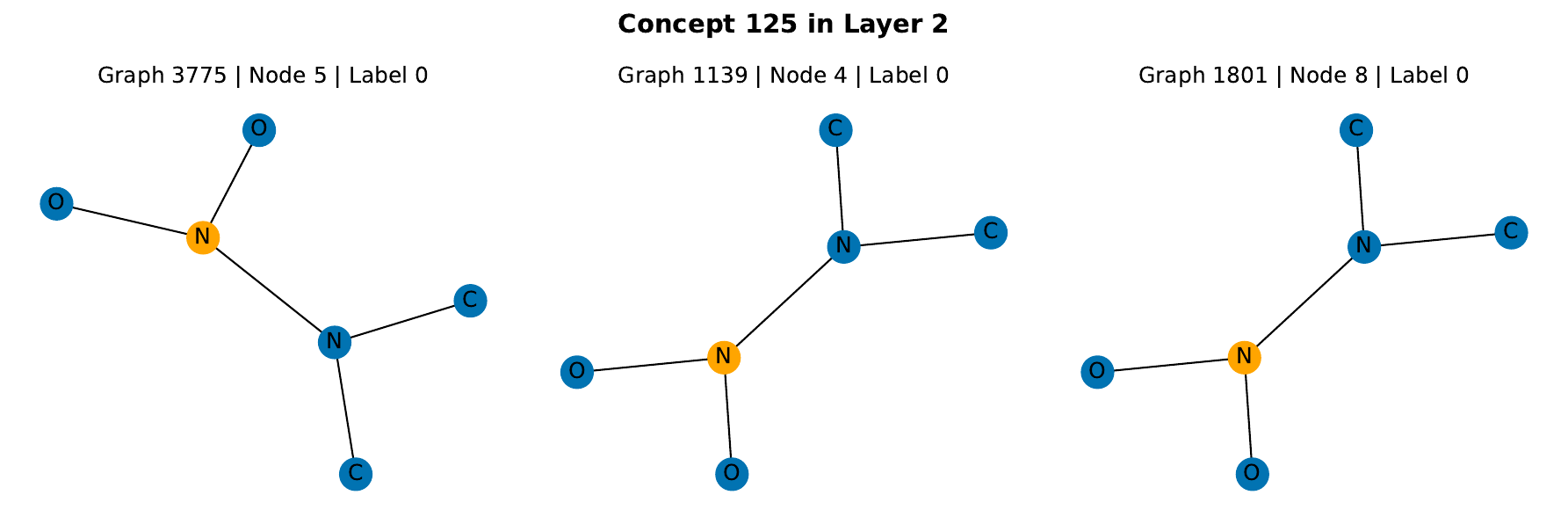}}
\caption{A concept discovered in the second layer of a model using Concept Graph Convolution (CGC) layers, trained on the Mutagenicity dataset. The concept identifies the desirable $NO_2$ structure, predictive for the mutagenic class. The orange nodes represent the nodes belonging to the concept, while the blue nodes are the expanded neighbourhood. Node labels correspond to the atom label.}
\label{cgc_mutag_concept2}
\end{figure}

Another benefit of extracting concepts at different layers is that earlier layers aggregate a smaller receptive field, wherefore, the $p$-hop neighbourhood visualized for the concept is smaller. The concepts extracted at earlier layers can potentially be easier to reason about, because of their smaller size, however, they may not present the full picture. For example, Figure \ref{cgc_mutag_concept1} shows a concept discovered for the Mutagenicity dataset at layer 2, showing a clearly repeated substructure across different molecules. Notice, that this seems to be a concept shared across mutagenic and non-mutagenic molecules, since these subgraphs are found in graphs of both classes. Furthermore, we are able to discover the desired set of concepts with the CGC. For example, Figure \ref{cgc_mutag_concept2} shows the concept identifying the $NO_2$ structure, defined as a desirable motif for the dataset for predicting molecules as mutagenic. Notice, that we discover the motif in the second layer of a model with three layers. We do not find this structure represented in the concept discovered in the third layer, which may be explained by the receptive field being larger, drowning out the contribution of the small structure. This highlights that it is desirable to understand the GNN at different stages of the message passing process, as it identifies different substructures as a cluster, representing a concept. Overall, concept distillation at every neighbourhood aggregation step of the model is required for a holistic understanding of the model and its focus at different layers, similar to convolutional layers in CNNs being described as learning specific features important for prediction via filters \citep{class_activation_mapping, gradcam, farvardin2026end, yang2025interactive}.
 
\begin{figure}[h]
\centerline{\includegraphics[width=1\textwidth]{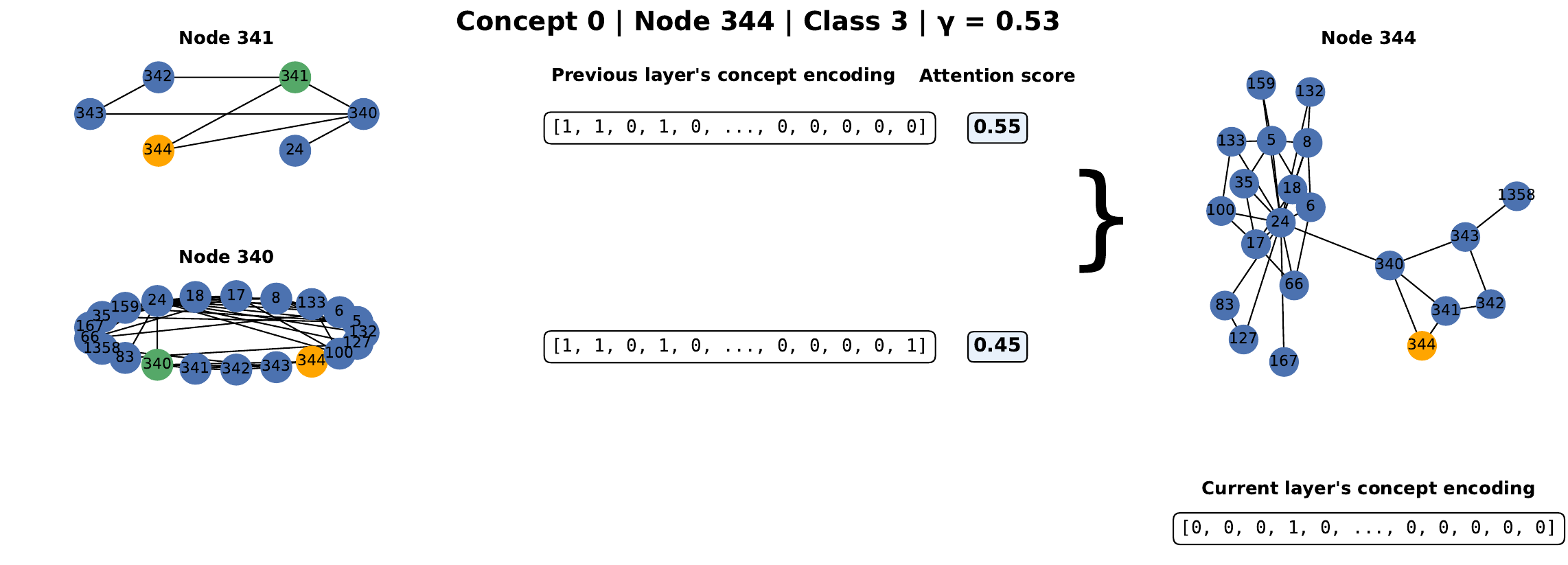}}
\caption{A concept discovered for the BA-Community dataset at the third pure Concept Graph Convolution (CGC) layer of the model. The graph on the left visualises the concept discovered, while the graphs on the right are the expanded graphs of the adjacent nodes in the previous layer. The orange node is the node assigned to the concept, while the green nodes are the respective neighbour in question and the blue nodes represent the other neighbours.}
\label{att_fig2}
\end{figure}

\begin{figure}[h]
\centerline{\includegraphics[width=1\textwidth]{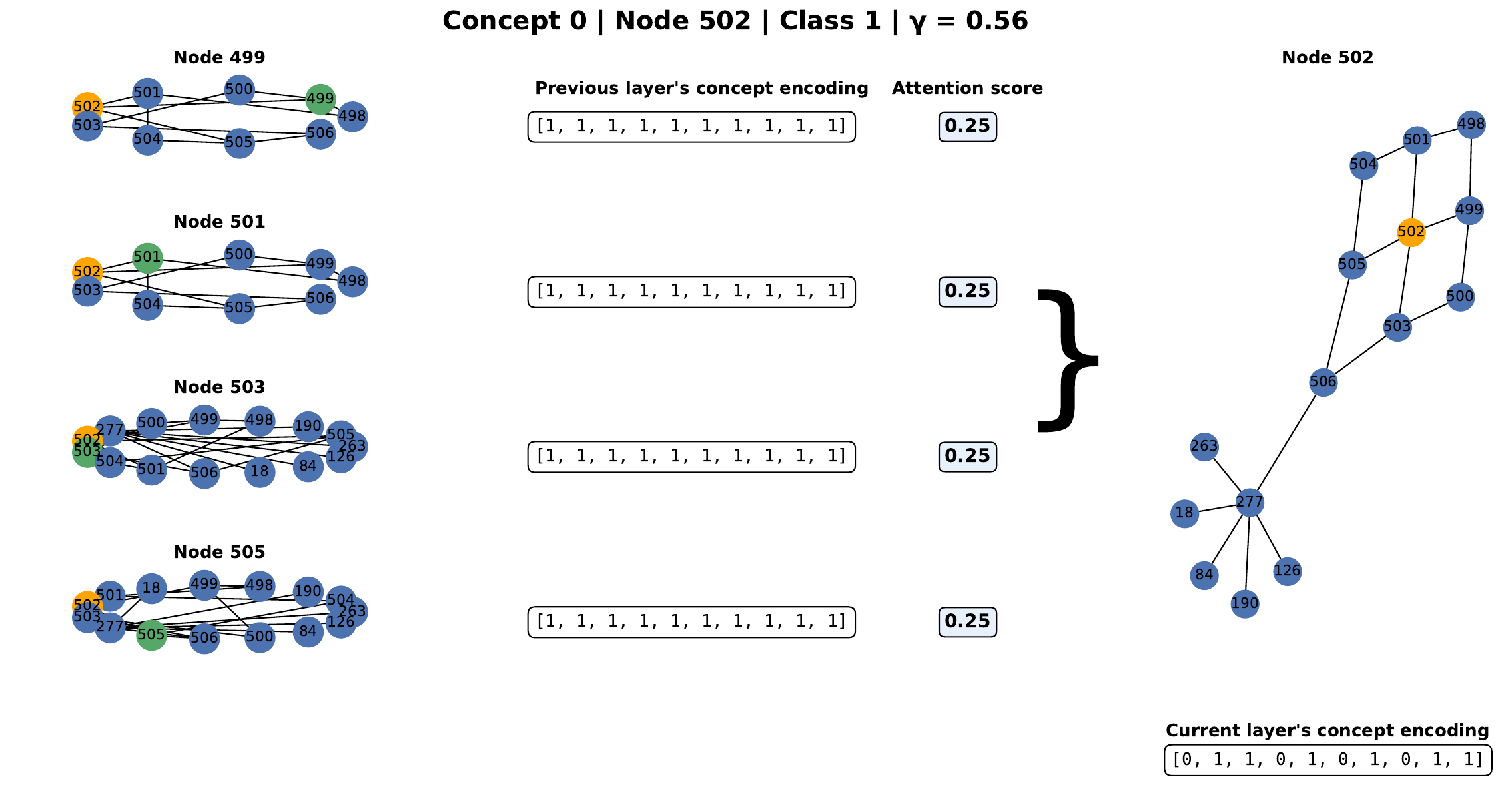}}
\caption{A concept discovered for the BA-Grid dataset at the second Concept Graph Convolution (CGC) layer of the model. The graph on the left visualises the concept discovered, while the graphs on the right are the expanded graphs of the adjacent nodes in the previous layer. The orange node is the node assigned to the concept, while the green nodes are the respective neighbour in question and the blue nodes represent the other neighbours.}
\label{att_fig3}
\end{figure}

Despite the low $\gamma$ values, we observe attention scores aligned with ground-truth structural importance across the other datasets. For example, Figure \ref{att_fig2} shows a concept distilled in layer 3, representing the top node of a house structure in the BA-Community dataset. The subgraphs displayed to the left in the figure, highlight that the important nodes identified by attention are the two middle nodes of the house structure. Both nodes receive a similar attention value of 0.55 and 0.45. The node receiving an attention score of only 0.45 is the node attached to the base structure. This makes intuitive sense, as the other middle of the house node is a cleaner representation of the house structure, while the node also attaching to the base graph is influenced by the representations of the nodes not belonging to the house structure. Examining the concept encoding vector, we can see that the encoding is slightly different, with the last value being 0 for the node only attached to other house-structure nodes. In contrast, it is 1 for the node also attaching to the base graph. This shows that the two nodes are close in the concept latent space, but have slightly different encodings due to the difference in neighbours, leading to slightly different attention scores. We would like to note though, that we do not find the positions of the concept encoding vectors to encode any further conceptual meaning, but simply identify the location of the cluster in the concept latent space. In contrast, Figure \ref{att_fig3} explains the computation of a concept representing the middle node of the grid structure in the BA-Grid dataset. Here, all nodes receive the same amount of attention, which can be explained by the grid structure being symmetrical across this middle node. When we examine the concept encoding vectors, we find that all nodes are assigned the same concept encoding vector. This can be attributed to each of the nodes being the same node in the graph structure, namely the middle node on the outside of the grid. Examining the attention scores in relation to the $\gamma$ value of 0.56, we can interpret the representations to be equally influenced by structure and concept attention as they encode the same information, due to the symmetry of the grid structure. The benefit of the structural edge weights here, is that they account for one of the corner nodes in the grid structure being attached to the base graph, influencing the neighbouring middle nodes. This analysis of the concepts and their computation is enabled by the concept-focused architecture of CGC layer, providing an increased insight into the reasoning of GNNs and what is important for them.

\section{Comparison to relational concept bottleneck models}
\label{rcbms}

We compare our method to R-CBMs \citep{barbiero2024relational}, because R-CBMs also perform message passing in the concept space. Table \ref{fig:rcbms_compare} summarizes the predictive accuracy of GNN models using CGC and pure CGC layers versus R-CBMs. Overall, R-CBMs exhibit better predictive accuracy than models using CGC and pure CGC layers. Most notably, R-CBMs achieve an accuracy of 92.2\% on the BA-Community dataset, while a GNN using pure CGC layers only achieves an accuracy of 82.69\%. However, the insight gained into the model is different. In the node classification case, the node features inputted to R-CBMs are initialized with the embeddings of two GCN \citep{Kipf2016GCNConv} layers. This means that the concept atoms used are the outputs of the GCN layer, which we consider not interpretable. In contrast, the CGC aims to replace GCN layers and directly operates in the concept space. This highlights a core difference between the proposed graph convolution and R-CBMs. CGCs aim to make message passing itself more interpretable. In contrast, R-CBMs are a variant of concept bottleneck models, which only aim to make a subsection of the model more interpretable. Moreover, the explanations extracted from R-CBMs are logic sentences combining the concept atoms. In contrast, the explanations extracted from CGC layers are intermediate node-level concept representations, wherefore, we do not compare them.

\begin{table*}[ht]
\centering
\resizebox{0.9\textwidth}{!}{
\begin{tabular}{llll}
\hline & \multicolumn{3}{c}{\textbf{\begin{tabular}[c]{@{}c@{}}Model Accuracy (\%)\end{tabular}}} \\
                        & \multicolumn{1}{c}{\textbf{CGC}}                       & \multicolumn{1}{c}{\textbf{Pure CGC}} & \multicolumn{1}{c}{\begin{tabular}[c]{@{}c@{}}\textbf{R-CBMs}\end{tabular}} \\ \hline
\textbf{BA-Shapes}   & 98.86 (98.37, 99.34) & 98.71 (97.56, 99.87) & \textbf{100.00 (100.00, 100.00)} \\
\textbf{BA-Community} & 69.56 (66.17, 72.95) & 82.69 (77.51, 87.88) & \textbf{92.2 (88.92, 95.51)} \\
\textbf{BA-Grid} & 99.90 (99.63, 100.00) & 99.90 (99.63, 100.00) & \textbf{100.00 (100.00, 100.00)} \\
\textbf{Tree-Cycle} & 97.69 (95.17, 100.00) & 96.98 (93.80, 100.00) & \textbf{100.00 (100.00, 100.00)} \\
\textbf{Tree-Grid} & 99.43 (98.59, 100.00) & 99.19 (98.25, 100.00) & \textbf{99.92 (99.69, 100.00)} \\
\hline
\end{tabular}%
}
\caption{Model accuracy for graph neural networks using Concept Graph Convolution (CGC) and pure CGC layers compared to Relational Concept Bottleneck Models (R-CBMs, \citep{barbiero2024relational}).}
\label{fig:rcbms_compare}
\end{table*}


\end{document}